%% file: ursa_iclr2026malgai.tex
\documentclass{article} 

\input{sections/preamble.tex}

\begin{document}

\maketitle

\input{sections/abstract.tex}

\input{sections/introduction.tex}
\input{sections/related_work.tex}
\input{sections/architecture.tex}
\input{sections/experiments.tex}
\input{sections/discussion.tex}

\bibliography{general_references,scientific_agents}
\bibliographystyle{iclr2026_conference}

\input{sections/appendix.tex}

\end{document}

%% file: sections/preamble.tex
\usepackage{iclr2026_malgai,times}

\input{math_commands.tex}

\usepackage{hyperref}
\usepackage{url}
\usepackage{placeins}

\usepackage{graphicx}
\usepackage{algorithm}
\usepackage{algorithmic}

\usepackage{listings}
\usepackage{xcolor}
\usepackage{tcolorbox}
\usepackage{fancyvrb}
\usepackage{amsmath,amssymb}
\usepackage[x11names]{xcolor} %
\usepackage{wrapfig}

\floatstyle{ruled}
\newfloat{code}{thp}{lop}
\floatname{code}{Code Block}

\newcommand{\quotegray}[1]{{\leavevmode\color{DarkSlateGray4}#1}}

\newif\ifblindreview
\blindreviewfalse 

\newcommand{\institutionphrase}{%
  \ifblindreview
    [\emph{institution hidden for blind review}]%
  \else
    Los Alamos National Laboratory%
  \fi
}

\newcommand{\codereleasenote}{%
  \ifblindreview
    [\emph{public code hidden for blind review}]%
  \else
    \url{https://github.com/lanl/ursa}%
  \fi
}

\title{URSA: The Universal Research and Scientific Agent}

\author{
Michael Grosskopf, \And
Nathan DeBardeleben,  \And
Russell Bent,  \And
Rahul Somasundaram, \And
Isaac Michaud,  \And
Arthur Lui,  \And
Alexius Wadell, \And
Warren D. Graham,  \And
Golo A. Wimmer,  \And
Sachin Shivakumar, \And 
Joan Vendrell Gallart,  \And
Harsha Nagarajan,  \And
Earl Lawrence 
\thanks{ Use footnote for providing further information
about author (webpage, alternative address)---\emph{not} for acknowledging
funding agencies.  Funding acknowledgements go at the end of the paper.} \\
ArtIMis\\
Los Alamos National Laboratory\\
Los Alamos, NM 87545, USA \\
\texttt{mikegros@lanl.gov} \\
}

\newcommand{\quotes}[1]{``#1''}

%% file: math_commands.tex
\usepackage{amsmath,amsfonts,bm}

\def\eqref#1{equation~\ref{#1}}

\def\1{\bm{1}}

\DeclareMathAlphabet{\mathsfit}{\encodingdefault}{\sfdefault}{m}{sl}
\SetMathAlphabet{\mathsfit}{bold}{\encodingdefault}{\sfdefault}{bx}{n}



%% file: sections/abstract.tex
\begin{abstract}
Large language models (LLMs) have moved far beyond their initial form as
simple chatbots, now carrying out complex reasoning, planning, writing, coding,
and research tasks. These skills overlap significantly with those that human
scientists use day-to-day to solve complex problems that drive the cutting edge
of research. Using LLMs in \quotes{agentic} AI has the potential to
revolutionize modern science and remove bottlenecks to progress. In this work,
we present URSA, a scientific agent ecosystem for accelerating research tasks.
URSA consists of a set of modular agents and tools, including coupling to
advanced physics simulation codes, that can be combined to address scientific
problems of varied complexity and impact. This work highlights the architecture
of URSA, as well as examples that highlight the potential of the system.
\end{abstract}

%% file: sections/introduction.tex
\section{Introduction}
The promise of AI for accelerating science has quickly turned from a far-off
vision to a near-term reality for advancing cutting-edge research. Emergent
reasoning and planning capabilities in large language models (LLM) have opened
new avenues for automating complex science and engineering tasks and
eliminating human-driven bottlenecks in the research process. As an example,
consider scientific domains like inertial confinement fusion (ICF) and
materials modeling. These models rely on physics simulations to explore
hypothesis spaces and guide experimental research (in other words, a
language model is unlikely to just ``reason through'' a good design
without access to verifier tools). 
Unfortunately, these
high-fidelity simulations can take hours or even days on some of the largest
supercomputers, often delaying scientific discoveries by months or longer. A
major contributor to these inefficiencies is the prevalence of unproductive
simulations that fail to advance knowledge, presenting an opportunity for AI to
accelerate progress through better identification of simulations to run.

Recent progress in large-scale foundation models and autonomous
\quotes{agentic} tool use (e.g., code-generating assistants and
planner-executor architectures) suggests a path toward AI systems that can
process and reason over large amounts of data to decide what to simulate and
adapt hypotheses on the fly.  However, most demonstrations to date target
internet-scale tasks such as software debugging or web search, leaving open the
question of how an agent can combine successes in reasoning and coding tasks
with high performance scientific computing for high-impact, high-consequence
scientific applications.

In this work we present the URSA agentic workflow for accelerating scientific
efforts developed at 
\institutionphrase\
for use by the scientific
community. URSA uses a set of modular, composable agents coupled with tool use
to hypothesize, plan, and execute research tasks to supplement domain expert
expertise in accelerating research outcomes. URSA uses agents dedicated to
planning, hypothesizing, researching, and executing computational tasks. Some
agents can also leverage advanced scientific simulation such as trusted
radiation-hydrodynamics models for ICF simulation. 

\subsection{Contributions}

\begin{enumerate}
  \item \textbf{Agentic architecture for scientific tool use:} We introduce URSA
  which combines large-scale language-model planning, autonomous research, and
  LLM-driven design optimization. 
  \item \textbf{Composable agents for varying complexity:} Our approach builds on
  the recent success of Sakana \citep{lu2024ai} and similar workflows, introducing
  an architecture to generalize prescribed, linear processes by incorporating
  structures that support loops and other feedback mechanisms.  \item
  \textbf{Demonstration:}  We present a series of experiments to demonstrate the
  capability of URSA on increasingly complex problems.  \item \textbf{Leveraging
  physics simulation for design automation} We present results for a key
  component of simulation-based scientific discovery--utilizing computational
  models to identify promising designs. We show that URSA outperforms standard
  methods (Bayesian optimization) for a design optimization task utilizing
  radiation hydrodynamics simulation.
\end{enumerate}

Our study charts a concrete path toward AI systems that actively \textit{does}
science, while illuminating gaps that remain in the autonomous use of agents
for critical science applications.

%% file: sections/related_work.tex
\section{Related Work}
The literature on agentic AI is broad. Two recent surveys,
\citep{gridach2025agentic} and \citep{ren2025towards}, provide a good overview of
recent progress. Both provide a useful categorization of the components of such
a system. \citep{gridach2025agentic} divides the agentic discovery process into
ideation; experimental design and execution; data analysis and interpretation;
and paper writing and dissemination. They also examine implementation and
application datasets. \citep{ren2025towards} breaks out the key components of a
system in a planner, memory, and tool sets for execution. They further break
these down and cover the literature on each, including scientific application
domains. Similarly, \citep{zhouai} reviews several of the top approaches in the
context of scientific AI agents.

The Sakana AI Scientist papers \citep{lu2024ai, yamada2025ai} are influential
exemplars of these end-to-end approaches. This work is the most closely related
to what we present here. These represent an attempt to build an end-to-end
automated approach to machine learning research. The system generates ideas,
reviews them for novelty, and scores them. The selected idea is implemented
along with numerical experiments. Finally, the idea is written up and reviewed.
Version 2 expanded the system's capability by using tree-search for the
experimentation and a vision-language model to improve the figures in the
written papers. Both versions use base models without additional fine-tuning.
Our approach builds on these ideas by developing agents that support less
regimented workflows, such as loops that allow ideas to be approved by
feedback.

The Aviary system \citep{narayanan2024aviary} also represents a complete
approach, but focuses on training complete systems of agents with a
reinforcement learning approach to fine-tuning LLMs. They introduce the concept
of a language decision process and provide two approaches to training. They
also consider tool use and decision-making to support it. They find that small
LLMs can be trained to match the performance of larger frontier models.

Other notable recent examples include Google's Co-Scientist
\citep{gottweis2025towards} which includes a sophisticated hypothesis and
planning agent, SciAgents \citep{ghafarollahi2024sciagents} which uses a
knowledge graph to build hypotheses from disparate scientific domains, and the
Agent Laboratory \citep{schmidgall2025agent} which considers a complete
framework. Additionally OpenAI's Deep Research\citep{deepresearch} does complex
research and formats a thorough report with detailed citations for a given
topic.

%% file: sections/architecture.tex
\section{Architecture}
The URSA agentic workflow is built on a set of specialized agents that are
composable for solving complex problems. It is able to hypothesize about
potential solutions to a problem, utilize web search and scraping to process
information from the internet, build a project plan for solving a given
problem, and perform tool-calling actions to solve the problem. Each agent
consists of a hybrid of explicit coding instructions and prompts that are used
to define the function of the agent. These agents are constructed using
LangGraph \citep{langgraph} where LLMs are used as the backend technology. In
the following subsections, we outline each of these agents in more detail.
To support reproducibility and community use, URSA is released as an open-source
framework at \codereleasenote.

\subsection{Planning Agent}
\label{sec:planner}
The URSA planning agent decomposes an input problem into a structured sequence
of executable steps. Implemented as a LangGraph network, it consists of three
LLM-backed nodes: a plan generator, a reviewer, and a formalizer (see 
Figure~\ref{fig:planning_agent} and 
Code Block ~\ref{code:planning_agent} in Appendix).
The generator proposes a step-by-step plan, the reviewer iteratively
refines it for clarity, completeness, and feasibility, and the formalizer
converts the approved plan into a structured JSON representation for downstream
execution.

\begin{figure}
  \centering
  \begin{minipage}{\linewidth}
    \centering
    \includegraphics[width=0.5\textwidth]{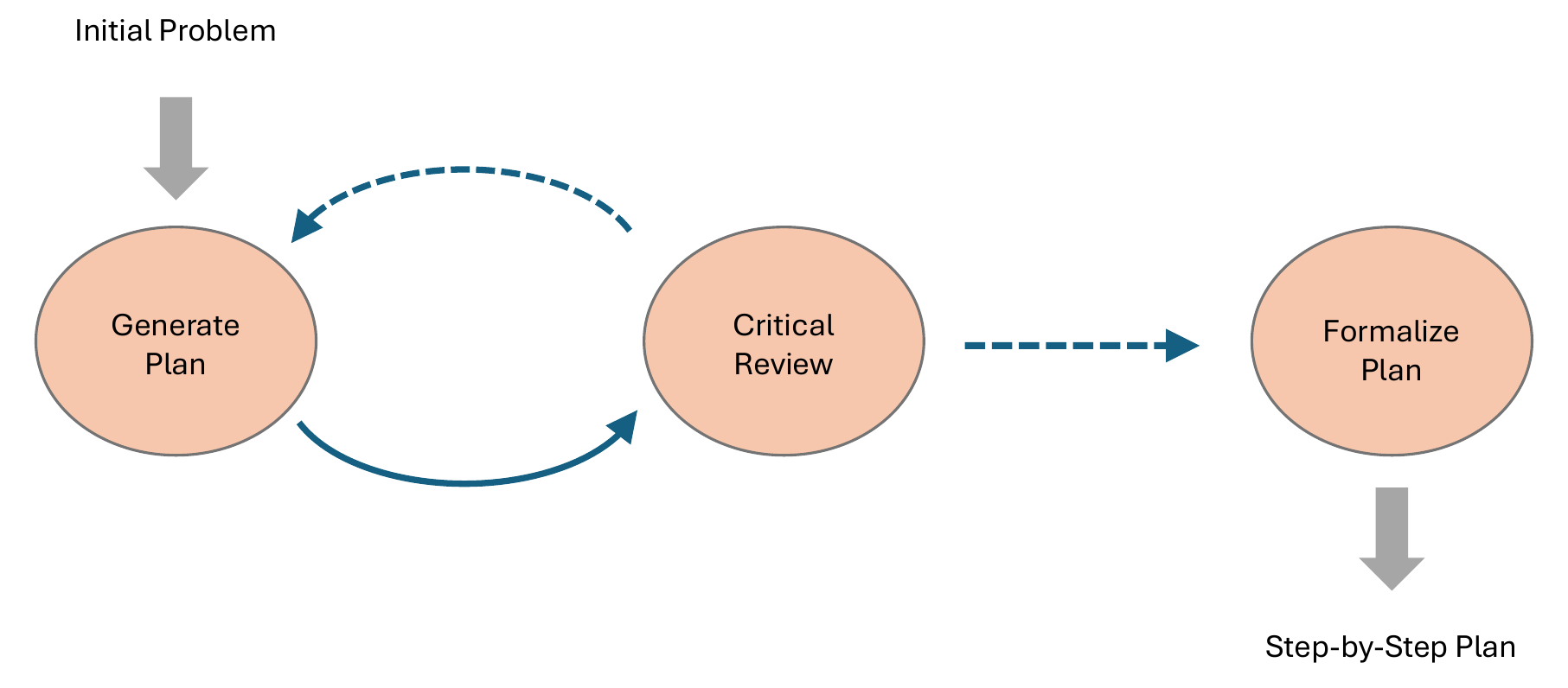}
  \end{minipage}
  \caption{Graphical workflow for the Planning Agent.}
  \label{fig:planning_agent}
\end{figure}

The proposed plan is iteratively reviewed for clarity, completeness, relevance,
and feasibility, and either approved or returned for revision.
Once approved, the plan is converted into a structured JSON representation
specifying each step’s description, requirements, expected outputs, and success
criteria, enabling reliable downstream execution.  
If the response does not conform to the JSON specification, the response and an
error message are appended to the prompt and provided back to the LLM. These
steps are repeated $f_{max}$ times before terminating with an error.  Details
of the planner prompt are provided in the supplemental material of the
appendix.

\subsection{Execution Agent}
\label{sec:executor}

\begin{figure}[h]
  \centering
  \begin{minipage}{\linewidth}
    \centering
    \includegraphics[width=0.75\textwidth]{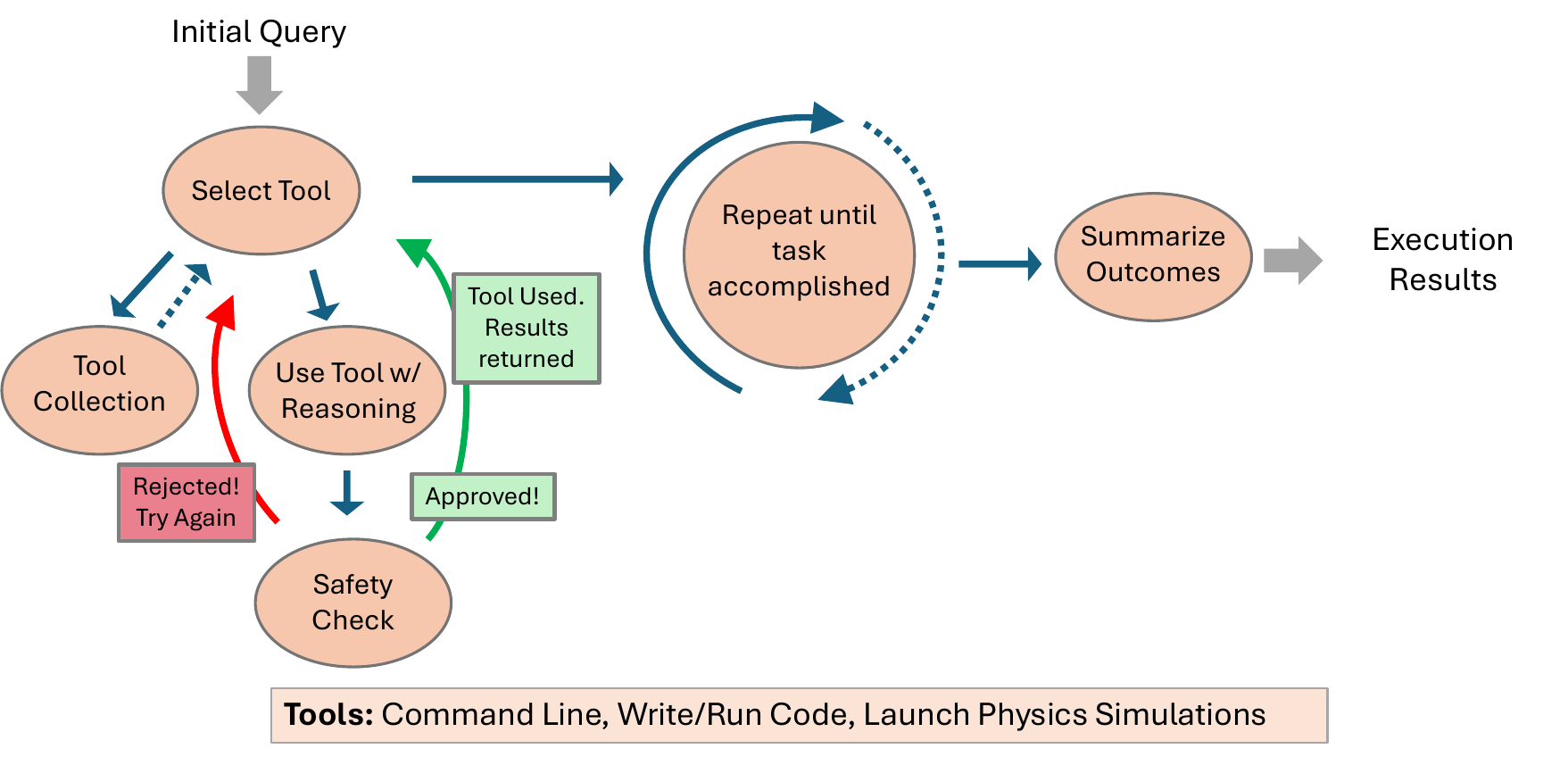}
  \end{minipage}
  \caption{Graphical workflow for the Execution Agent.}
  \label{fig:exec_agent}
\end{figure}

The URSA execution agent carries out code and tool-using tasks to perform steps
necessary to solve a given problem. The agent is passed a general problem
prompt or a particular step as part of a larger plan. These actions are carried
out through use of tool calls (LangGraph) as well as use of MCP servers (including
remotely hosted), as shown in
Figure~\ref{fig:exec_agent}.
This allows the LLM to autonomously select
the appropriate tool for a given task and iterate on the tool to diagnose and
fix problems. 
After the execution agent has finished using the available tools to
solve the problem, the 
results are summarized for the user or for downstream communication in more
complex workflows.

\begin{figure}
  \centering
  \begin{minipage}{\linewidth}
    \centering
    \includegraphics[width=0.55\textwidth]{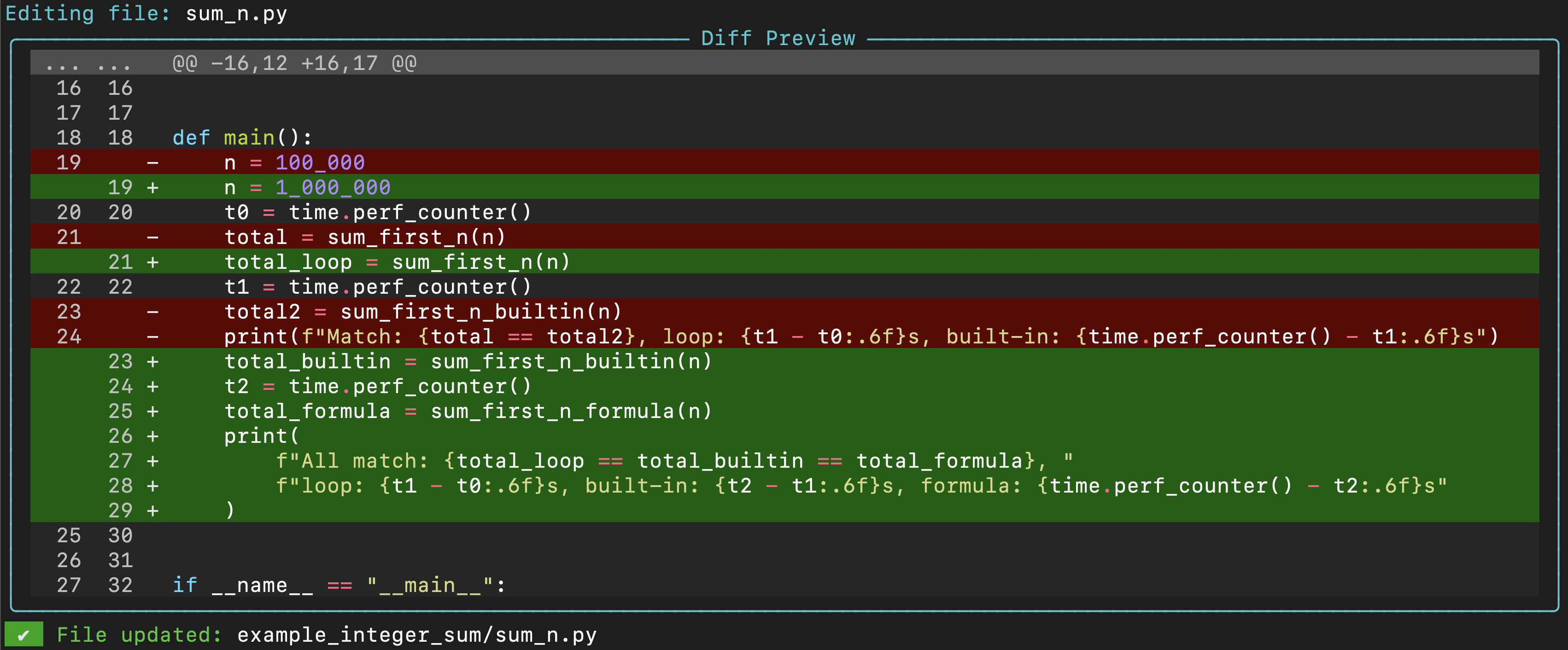}
  \end{minipage}
  \caption{Example of source code diffs sent by Execution Agent.}
  \label{fig:diff}
\end{figure}


The execution agent interacts with its environment through a small set of tools
for writing files and executing system commands (see 
Code Block ~\ref{code:execution_agent} in Appendix). 
Commands are
subject to a lightweight LLM-based safety check prior to execution, reducing
the risk of unintended actions. Figure~\ref{fig:diff} 
illustrates an example where the agent
iteratively modifies previously written files. Full execution prompts are
provided in the appendix.

\subsection{Research Agent}
\label{sec:researcher}
The URSA research agent utilizes web search and scrapes web content to collect
and summarize information for solving a given problem. Like the planning agent,
the research agent consists of a generation phase, a review phase, and a
summarization phase. The generation phase uses tool calls to a web search tool
or a web parser tool to gather information for addressing the problem 
(see Figure~\ref{fig:research_agent} and 
Code Block ~\ref{code:research_agent} in Appendix).
The web parser is given a URL and context by the
LLM. It uses the BeautifulSoup \citep{richardson2007beautiful} python package to
scrape information from the URL as text. Then, to avoid carrying excess tokens
downstream in the workflow and provide more compact information for later
processing, an LLM is invoked to summarize the text from the URL in the given
context.
This result is returned from the tool to the research agent.
The generation phase proposes an answer given the context of the tool calls,
which is reviewed for accuracy, completeness, and diligence. These two phases
are repeated until the reviewer approves the solution or a set maximum number
of iterations. The results are then summarized in the context of the original
query and returned to the user or sent downstream in the workflow.

\begin{figure}[h]
  \centering
  \begin{minipage}{\linewidth}
    \centering
    \includegraphics[width=0.5\textwidth]{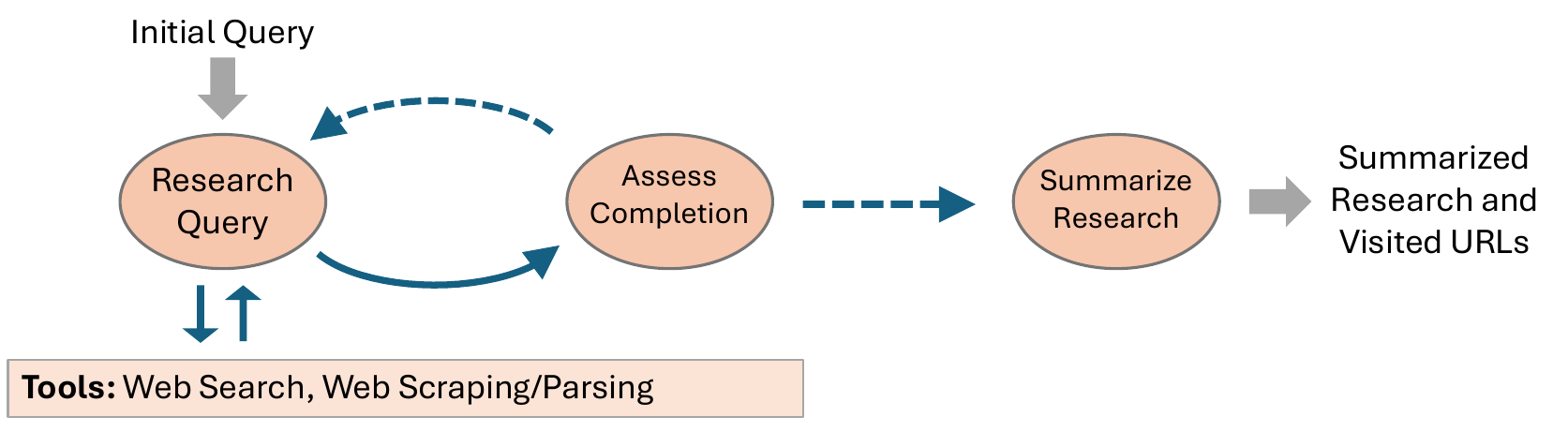}
  \end{minipage}

  \caption{Graphical workflow for the Research Agent.}
  \label{fig:research_agent}
\end{figure}

\subsection{Hypothesizer Agent}
\label{sec:hypothesizer}
\begin{figure}[h]
  \centering
  \begin{minipage}{\linewidth}
    \centering
    \includegraphics[width=0.5\textwidth]{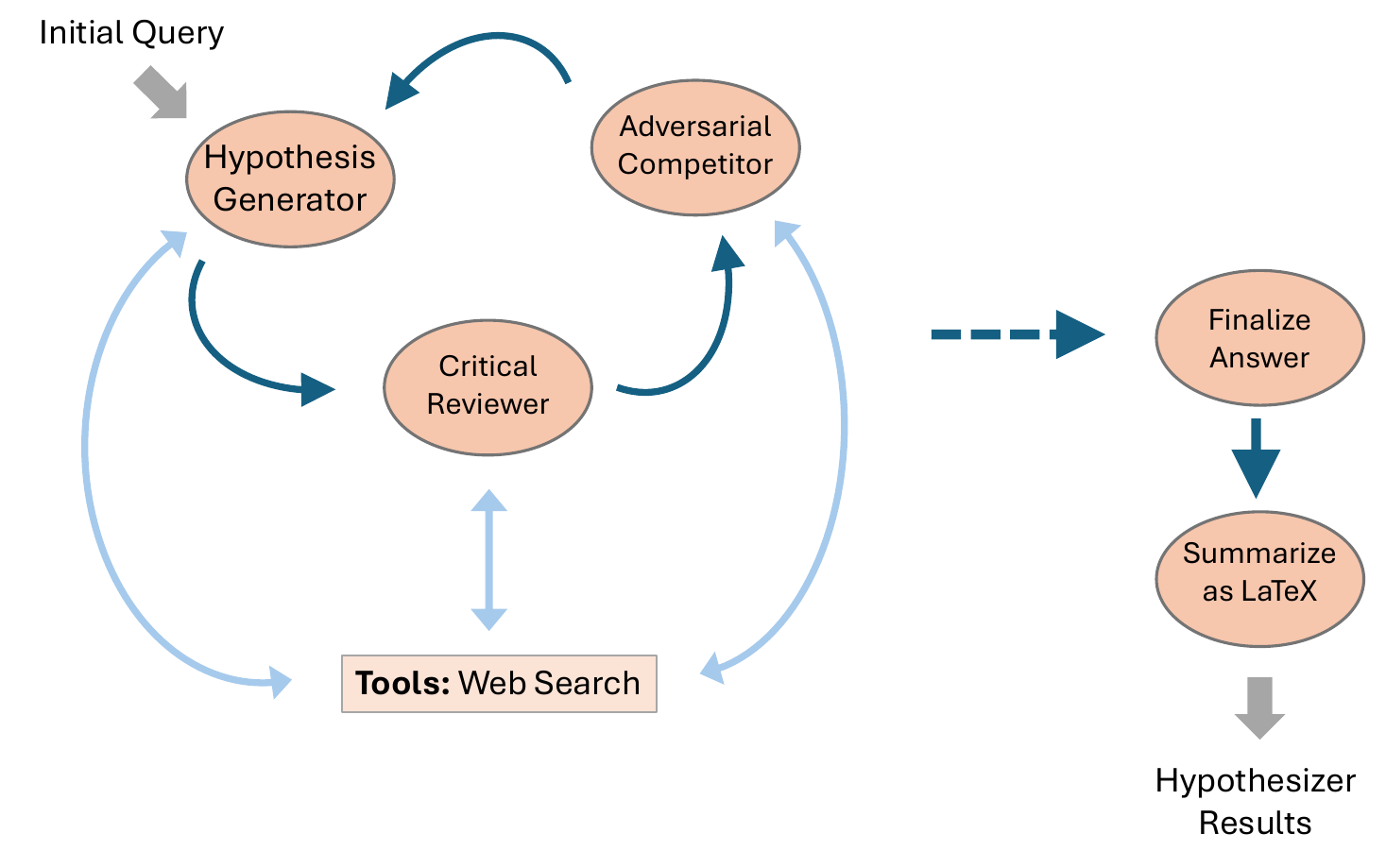}
  \end{minipage}

  \caption{Graphical workflow for the Hypothesizer Agent.}
  \label{fig:hypothesizer_agent}
\end{figure}

The goal of the URSA hypothesizer agent is to utilize web search and a vigorous
debate to hypothesize a solution to a user prompt. The difference between the
hypothesizer agent and the planning/research agents are an internal iteration
for solving the problem and the structure of output. The hypothesizer consists
of three internal subagents: the hypothesis generator, the critic, and the
competitor, as shown in Figure~\ref{fig:hypothesizer_agent}. 
The hypothesis generator performs a web search to generate
summaries of information available on the internet 
(see Code Block \ref{code:hypothesizer_agent} in Appendix). 

Unlike the research agent, it does not parse information directly from the
individual results but uses information summarized from the web search in its
generation. The initial hypothesis is then passed to the critic to identify
flaws or areas of improvement.
Both of these results are then passed
to the competitor who assimilate the critiques and propose an approach to
counter the initial hypothesis.
This feedback is given to the
hypothesis generation subagent to propose changes to the hypothesis.
This cycle is repeated until a maximum number of iterations, at which point the
complete debate is used to produce a complete solution to the initial query.
Details of the prompts for each LLM call are provided in the
supplemental material of the appendix.


\subsection{ArXiv Agent}
\label{sec:arxiv_agent}

\begin{figure}[h]
  \centering
  \begin{minipage}{\linewidth}
    \centering
    \includegraphics[width=0.5\textwidth]{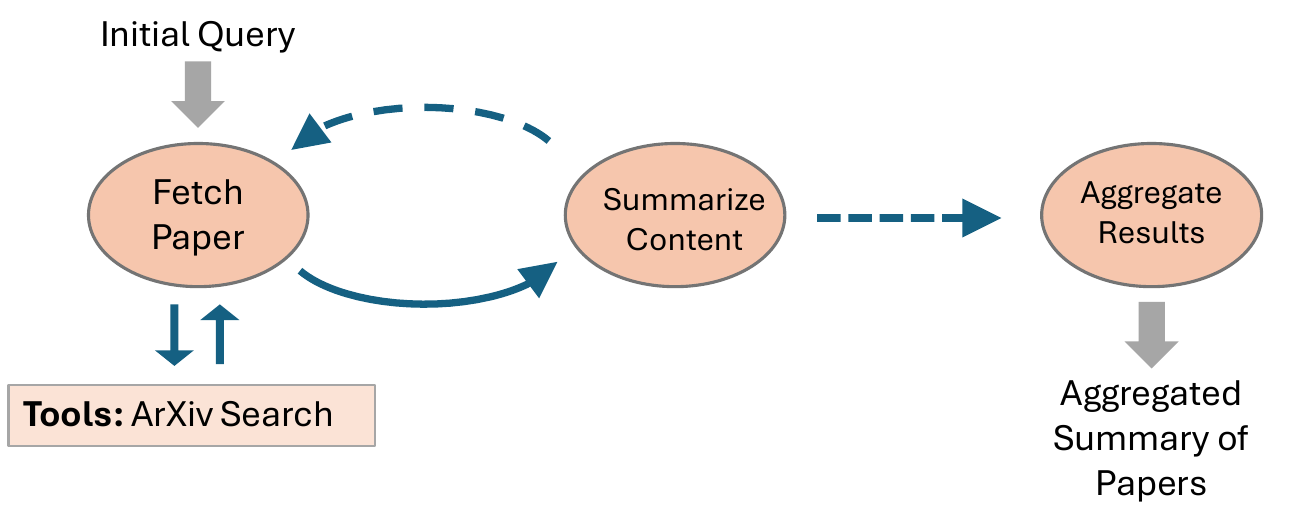}
  \end{minipage}

  \caption{Graphical workflow for the ArXiv Agent.}
  \label{fig:arxiv_agent}
\end{figure}

The e-print repository ArXiv provides an open access store of research prints
\citep{ginsparg1994first,ginsparg2011arxiv}. The goal of the URSA ArXiv agent 
(Figure~\ref{fig:arxiv_agent}) is
to utilize the ArXiv search API to find papers relevant to a given problem and
then use an LLM to process the text and images in the paper to summarize the
cutting-edge research related to the motivating problem.


Given a search query, the ArXiv agent retrieves a set of relevant papers using
the ArXiv API and extracts both textual content and figures. An LLM summarizes
each paper, incorporating brief descriptions of key images, and the individual
summaries are aggregated into a concise, context-aware overview of the
literature.  
See Code Block \ref{code:arxiv_agent} in the Appendix for more details.

Then, the full text of each paper, including the actual text and the image
descriptions, is fed into a \textsc{summarize node} which provide a summary of
the full text. 
The papers are processed independently in this manner
and each summary is aggregated to provide a detailed but concise overview of
the ArXiv literature on a particular topic in the given context of interest. 

In Appendix \ref{app:arxiv_example}, we show an example of using the ArXiv
Agent to provide a contextual summary on estimates of neutron star radii from 3
papers on the arXiv using \verb|o3|. While automated literature review is
directly useful to researchers, coupling this agent to other URSA agents either
in a workflow or as a tool unlocks the potential for agents to perform
on-the-fly research to assist in autonomous science and problem solving.

%% file: sections/experiments.tex
\section{Experiments}

To highlight the capability of URSA, we discuss a series of examples with
increasing complexity. Rather than a prescribed linear workflow, the agents in
URSA are deployed flexibly to solve basic problems that accelerate short term
tasks as well as complex tasks that take multiple planning steps or even
substeps. 

\subsection{6-Hump Camel Optimization}
To demonstrate a low-complexity example task, we used URSA to write code to
optimize the six-hump camel function, a common multi-modal test function for
demonstration of global optimization techniques\citep{molga2005test}. The
Execution Agent from Section \ref{sec:executor} was given the following prompt:
\begin{tcolorbox}[colback=blue!5!white] Optimize the six-hump camel function.
Start by evaluating that function at 10 locations.  Then utilize Bayesian
optimization to build a surrogate model and sequentially select points until
the function is optimized.  Carry out the optimization and report the results.
\end{tcolorbox}

\begin{wrapfigure}{r}{0.45\textwidth}
  \centering
  \vspace{0.0cm}
  \includegraphics[width=0.45\textwidth]{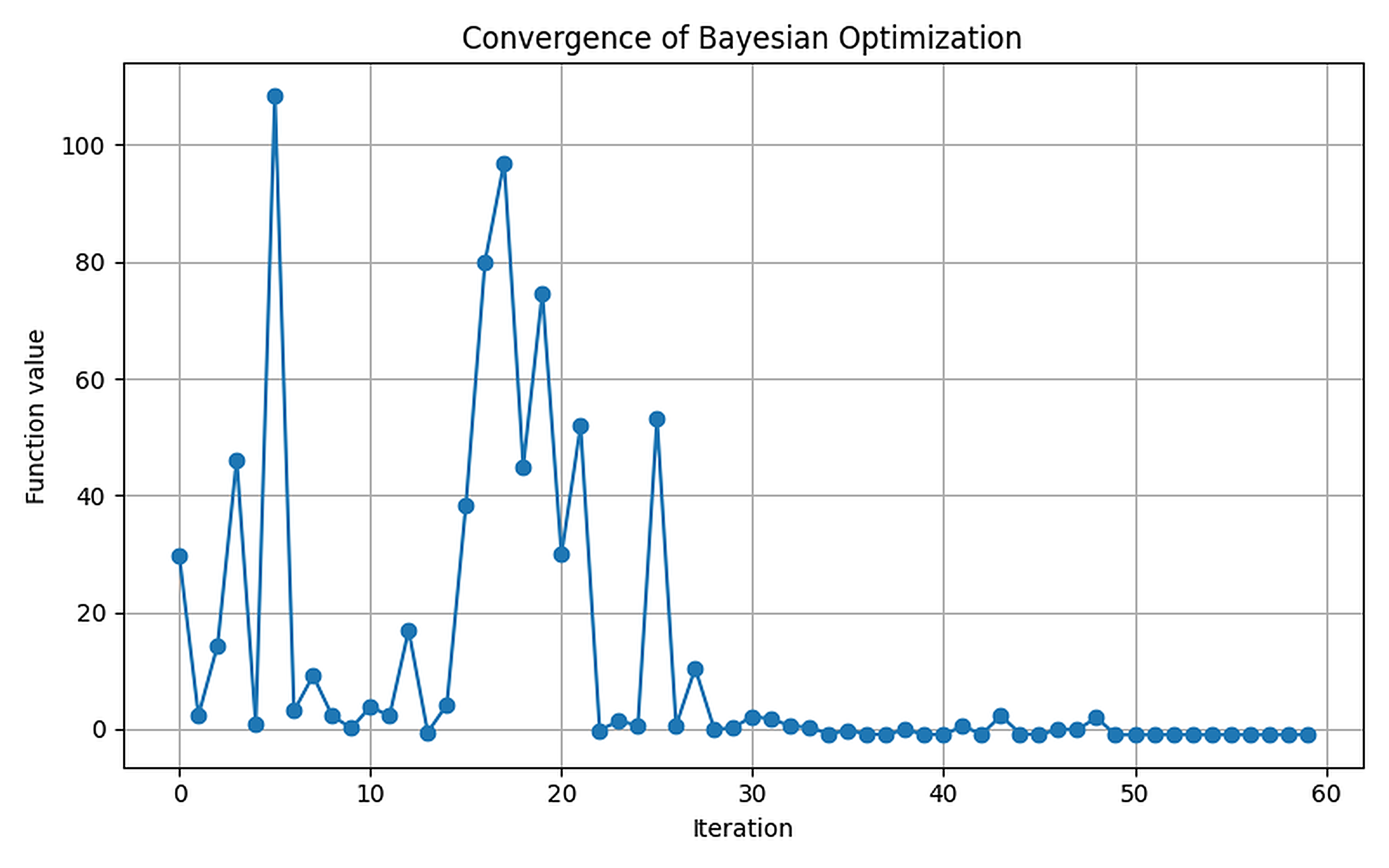}
  \caption{Convergence plot of the optimization of the six-hump camel function as generated by the URSA written and evaluated Bayesian optimization script.}
  \label{fig:BO}
  \vspace{-0.85cm}
\end{wrapfigure}


URSA's execution agent, with the OpenAI \texttt{o3-mini} model as the LLM, wrote
a python script to define (correctly) the six-hump camel function, used
\texttt{gp\_minimize} in the scikit-optimize \citep{scikit-learn} package to perform
Bayesian optimization. It made a convergence plot (Figure \ref{fig:BO}) of the
optimization showing successful convergence to the known global optimum. The
running, writing code, evaluation, and plotting took a few minutes with no
human feedback. The resulting code was saved in a local workspace for reuse or
extension to new problems of interest to a researcher.

\subsection{Surrogate Model Building and Benchmarking}
\label{sec:surrogate_building}
To demonstrate to the flexibility of URSA on complex problems, we highlight an
example where the Planning and Execution Agents are combined. Here, the
Planning Agent breaks the problem down into a set of compact steps that the
Execution agent addresses. This follows the observations that splitting complex
problems into many smaller, easily solvable tasks improves agentic workflows
\citep{wang2024evaluating,schneider2025generative, wang2024strategic}.

In this example, we use URSA to process a dataset, build two probabilistic
surrogate models \citep{gramacy2020surrogates}, compare their quality of fit
visually and quantitatively, and compare the quality of their uncertainty
quantification. The data supplied for the surrogate models was a set of 484
evaluations of the Helios radiation-hydrodynamics simulator \citep{helios}. The
goal of the surrogate was to predict the log (base 10) neutron yield from a set
of five geometry parameters. 

To solve this, we used URSA to build a workflow with Planning Agents and an
Execution agent prompted with:
\begin{tcolorbox}[colback=blue!5!white]
\noindent\textbf{Surrogate model prompt (abridged).}
Load \texttt{finished\_cases.csv}, train GP (gpytorch) and BNN (numpyro),
evaluate $R^2$ and coverage, and generate fit/UQ plots.
Full prompt in Appendix~\ref{app:prompt-exp-surrogate}.
\end{tcolorbox}

\begin{wrapfigure}{r}{0.45\textwidth}
  \centering
  \vspace{0.0cm}
  \includegraphics[width=0.45\textwidth]{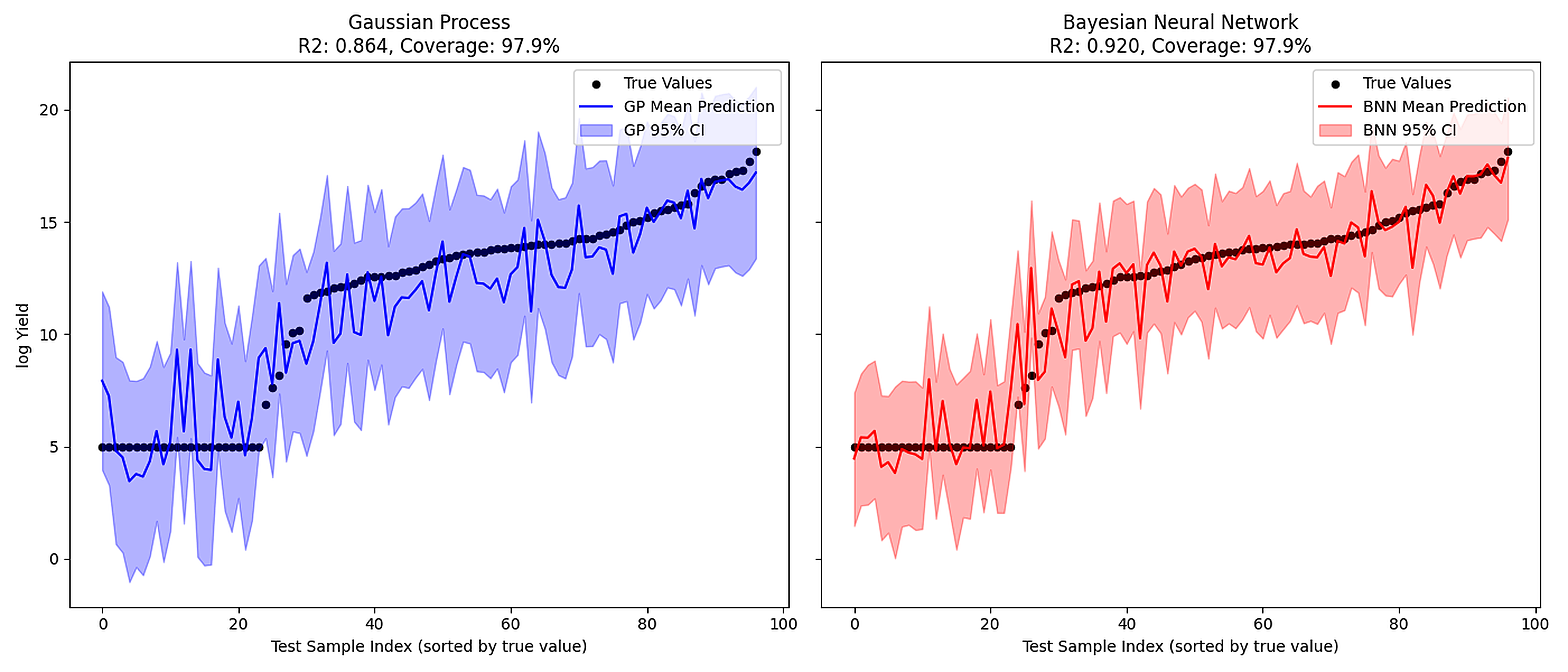}
  \caption{Prediction of log neutron yield in an ICF target from Helios
simulation using a Gaussian processes and Bayesian neural network. Both plots
were generated by URSA through the autonomous workflow, showing that the URSA
built surrogate models show strong predictive performance and uncertainty.}
  \label{fig:Surrogates}
  \vspace{-0.5cm}
\end{wrapfigure}

The Planning Agent decomposed the task into a structured multi-stage workflow
that was executed sequentially by the Execution Agent. The resulting pipeline
covered data validation and preprocessing, surrogate model fitting, predictive
evaluation, and uncertainty quantification.  To test robustness to
data-formatting errors, the prompt deliberately specified an imprecise target
column name, which the workflow resolved correctly.  The URSA-built surrogate
models were evaluated and plotted (autonomously) by URSA and are presented
in Figure~\ref{fig:Surrogates}.

\subsection{ICF Optimization With Helios}

Here, we demonstrate the capability for URSA to perform efficient design
optimization of a capsule for a double shell inertial confinement fusion (ICF)
experiment using the 1D radiation-hydrodynamics code Helios \citep{helios} as a
tool for the Execution Agent. Unlike traditional approaches to optimization,
agentic optimization leverages external data from literature, knowledge about
ICF trained into the LLM, and intelligent reasoning to identify designs to
evaluate. LLM-driven optimization has shown promising results for ML
hyperparameter optimization\citep{liu2024large}, a problem for which there is
abundant online information that can be trained into the model.

To perform the agentic autonomous ICF design, we used a combination of the URSA
hypothesizer agent and execution agents. The hypothesizer was given the
following prompt: 

\begin{tcolorbox}[colback=blue!5!white]
\noindent\textbf{Design prompt (abridged).}
Using prior indirect-drive double-shell ICF design principles
\citep{montgomery2018design}, plan and iteratively optimize a
direct-drive NIF experiment with a five-layer target
(Al ablator, foam cushion, Be tamper, Cr inner shell, DT fuel).
Evaluate candidate designs with the Helios radiation-hydrodynamics
model and maximize simulated neutron yield, targeting
$\log_{10}(\text{yield}) > 17$.
Full prompt in Appendix~\ref{app:prompt-exp-direct}.
\end{tcolorbox}

The Hypothesizer agent proposed the workflow described in Section
\ref{sec:hypothesizer} to come up with an initial design and passed that to the
execution agent from Section \ref{sec:executor}. 



\begin{wrapfigure}{r}{0.45\textwidth}
  \centering
  \vspace{0.0cm}
  \includegraphics[width=0.5\textwidth]{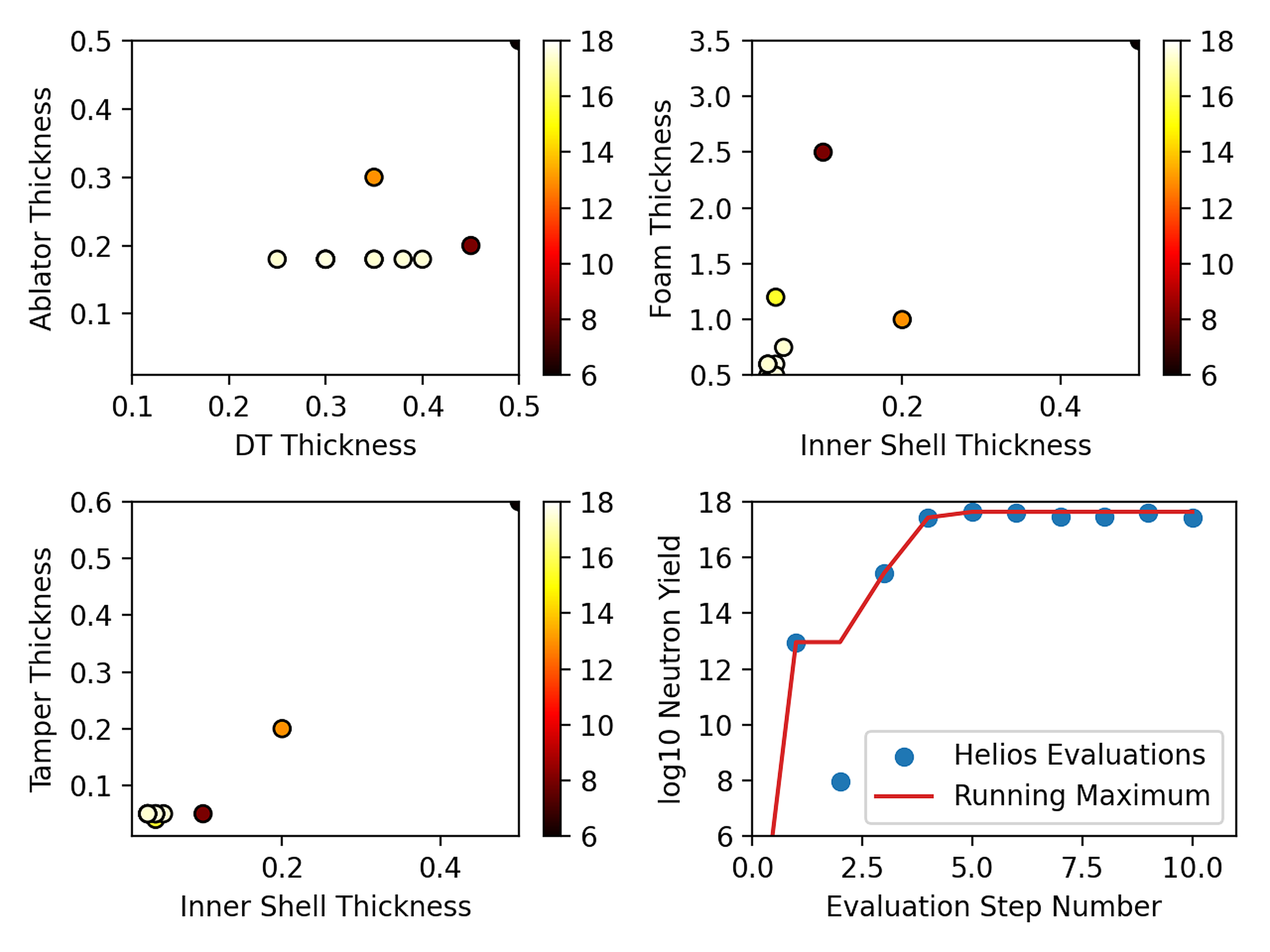}
  \caption{The sequence of evaluations of 1D Helios by the URSA Execution Agent
driven by o1. The right left shows the increasing performance over iterations,
while the other 3 plots show how the designs progressed through the parameter
space.}
  \label{fig:LLMDesignSequence}
  \vspace{-0.5cm}
\end{wrapfigure}

The Execution agent is then given the final summary and prompted ``Given that
plan, run Helios on a design that will generate a maximal yield.'' 
The
execution agent then proposed one or more designs to evaluate, reasoning about
improvements at each step. The execution agent was then prompted to continue
with ``Run Helios on a design that will generate an even higher yield'' ten
times. The final design was chosen as the highest performant design evaluated.

URSA, using \texttt{o3-mini} model for the hypothesizer and \texttt{o1} for the
executor, was able to identify near-optimal performant target geometries. In
Figure \ref{fig:LLMDesignSequence}, three plots show a bivariate projection of
the design space to indicate how the search winnowed in on a design region of
high performance. The first evaluated design is in the far upper right corner
of all three, which obtained no yield. The subsequent steps quickly move into
an area of high performance. The lower right panel shows the design performance
and running maximum yield with quick convergence to near optimal designs over
the specified $10^{17}$ threshold.

\begin{figure}[h]
  \centering
  \includegraphics[height=1.2in]{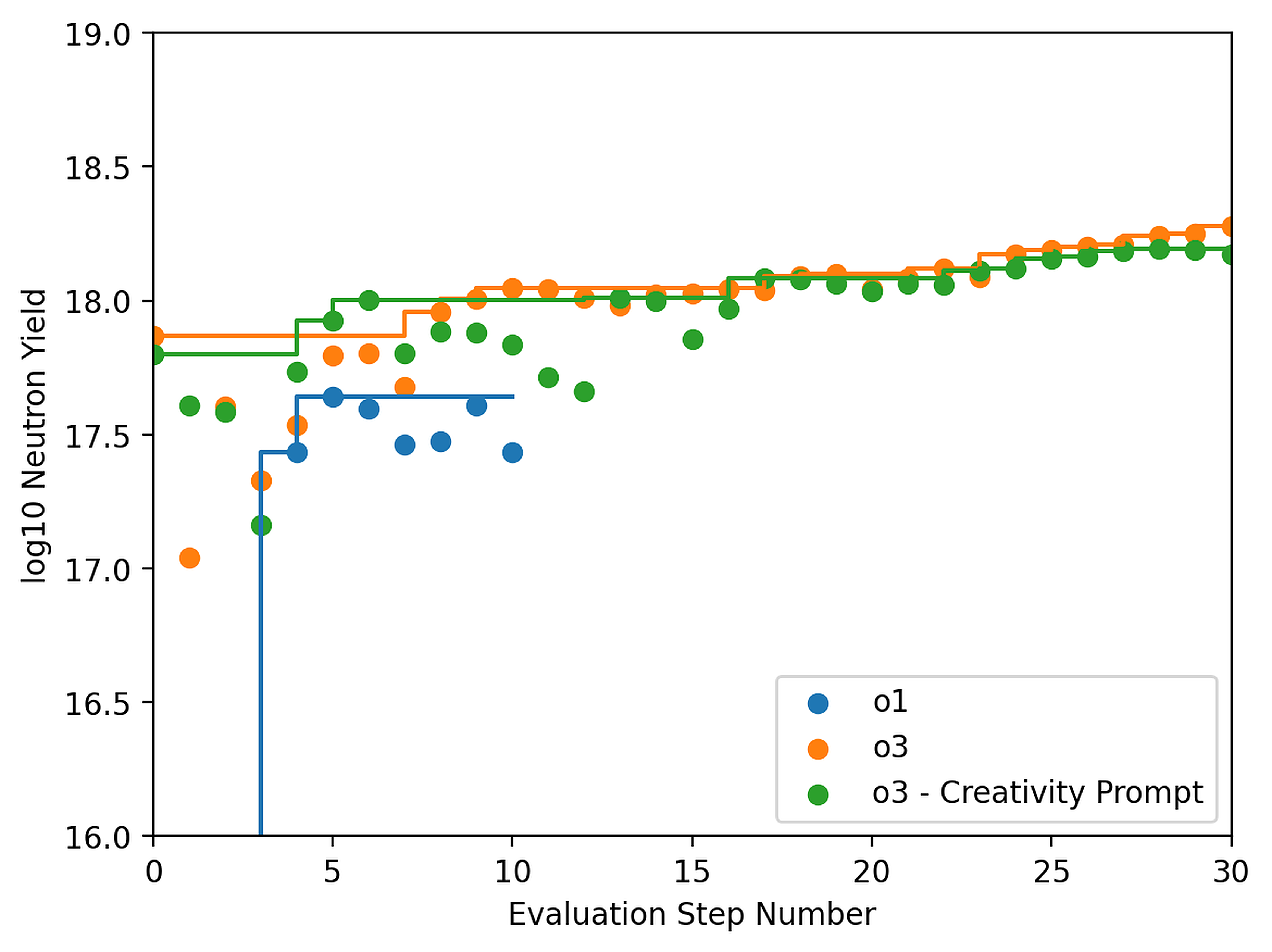}
  \includegraphics[height=1.2in]{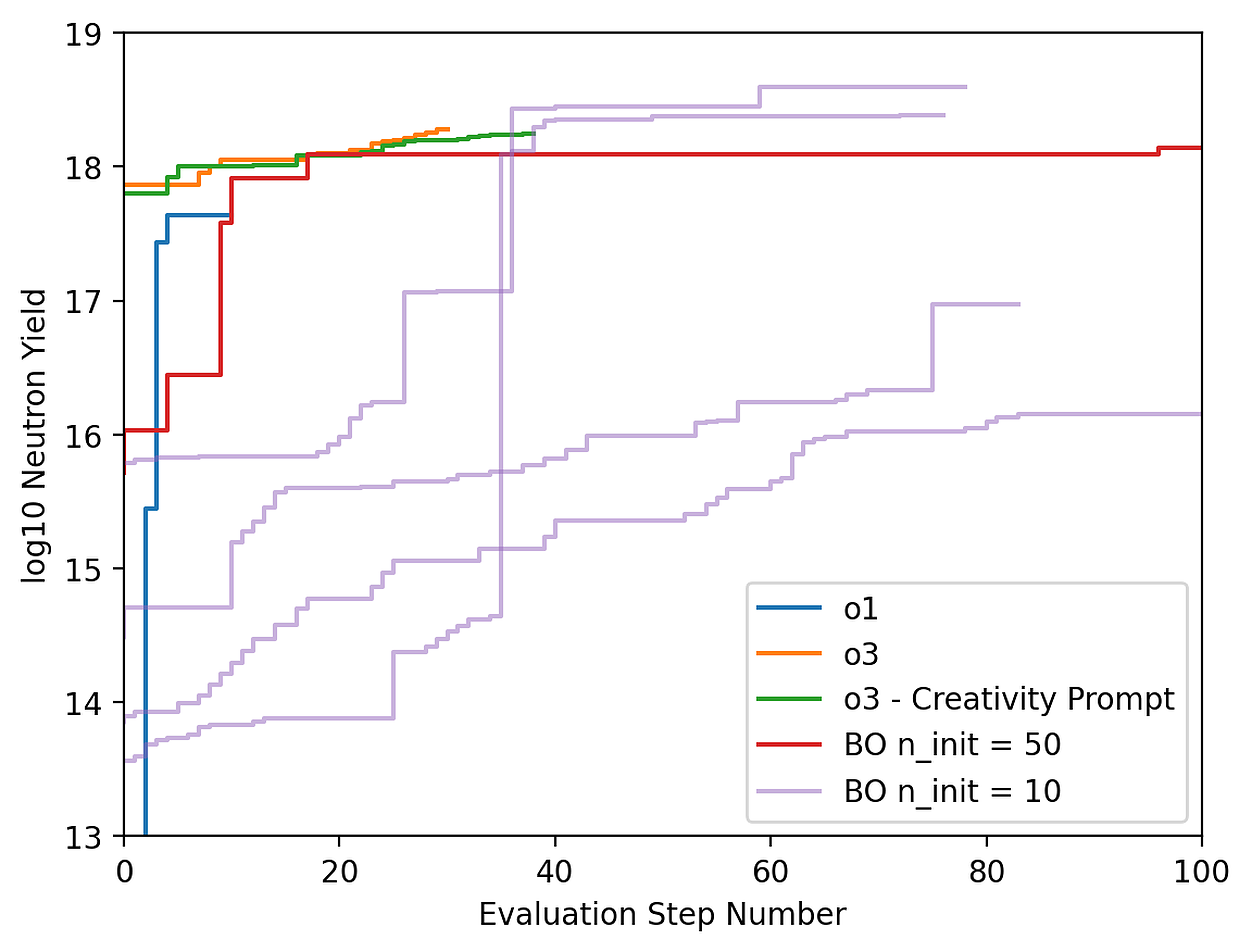}
  \caption{Comparison of URSA to Bayesian optimization for designing a
direct-drive ICF design. The plots show the running maximum neutron yield, with
the initial space-filling Latin hypercube random design for BO have been
removed to highlight only the cases where the data-driven BO model is compared
to the URSA literature-informed model. URSA was able to find near-optimal
designs faster and more reliably.}
  \label{fig:compare_BO}
\end{figure}

We also compare the design performance to two iterations of Bayesian
optimization (a standard approach for this problem)\citep{vazirani2021coupling},
one with 50 random space-filling points and four with 10 initial points (the
replicates for the smaller initial set were due to large expected variability
in performance from case to case). We also replicated the case from Figure
\ref{fig:LLMDesignSequence} using \texttt{o4-mini} 
for the hypothesizer and
\texttt{o3} for the executor. This was done twice, though an additional sentence
encouraging creativity was added in the second case 
(denoted \texttt{o3} - Creativty Prompt in Figure \ref{fig:compare_BO}). 
To reach comparable
performance to designs URSA found in under 10 model evaluations, Bayesian
optimization with $n_{init}=50$ required 18 additional runs for a total of 65
evaluations. For the $n_{init}=10$ case, the best required 37 runs for a total
of 47. URSA found near optimal designs in fewer evaluations than would be used
to initialize a Bayesian optimization loop.
We hypothesize that this is an example of LLM reasoning improving this search
operation over conventional Bayesian optimization.

%% file: sections/discussion.tex
\section{Discussion}


This work has developed an agentic workflow framework which builds on recent
successes using agentic AI to address scientific discovery applications. The
framework defines and implements concrete, generalizable agents that are
reusable and composable for a wide variety of uses, leveraging the strengths of
the latest LLMs under development, as demonstrated on several example problems.
These results yield several interesting future directions for this and other
frameworks to consider. First, the presented results experimented with OpenAI
models as the underlying technology used by the agents to execute the
workflows, and further results should evaluate alternative and hybrids of LLMs
for different agentic tasks, perhaps combined with fine-tuning to improve the
overall performance of the system. Second, the agentic workflows formalizes a
best practice when working with LLMs to address complex problems--breaking the
problem into small manageable pieces.  Further results should  experiment with
this concept to identify the degree to which underlying tasks should be
decomposed into individual agents.  Third, it would be interesting to implement
a parallelized, collaborative version of the workflows, where agents are given
the same task, compare results, and use each other's outcomes to inform future
actions.  Fourth, there remain a number of open questions related to ensuring
the fidelity of the results produced by AI agents.  Here, we expect future work
on fine-tuning LLMs, integrating alternate, non-LLM based AI models as the
underlying technology for some agents, and combining the agents with formal
methods for verification are all interesting future directions to address these
failures.

Beyond these directions, it is important to keep in mind limitations of the the
approach and broader potential for impact of this and similar agentic systems.
In Appendix \ref{ref:failures} we highlight a set of failure modes for URSA
which are important to be aware of. While hallucinations are a problem in all
human-LLM interactions, in long, complex workflows, hallucinations can be hard
to detect and invalidate all downstream results. 
URSA includes a checkpoint/restart capability that empowers a user to restart
at numerous points and ``steer'' the problem back on target.  This feature was
not utilized in the results presented in this paper, however.
It is important for generated
code results and data to be reproducible by a human and that the work actually
done in the agentic workflow can be clearly identified and logged. This becomes
even more important as frontier-class LLMs become more and more capable. LLMs
are already capable of generating convincing hallucinations and ensuring
logging of actions independent from the LLM will be increasingly critical for
building trust in agentic results.


%% file: sections/appendix.tex
\appendix
\section{Appendix}

\section{Agent Prompts}\label{app:prompts}
Here we document the prompts used for the different nodes and the different agents.

\subsection{Planning Agent Prompts}
\begin{tcolorbox}[colback=blue!5!white, colbacktitle=blue!40!white,title=\textbf{Planner Prompt}]
    You have been given a problem and must formulate a step-by-step plan to solve it.\\
    \\
    Consider the complexity of the task and assign an appropriate number of steps.
    Each step should be a well-defined task that can be implemented and evaluated.
    For each step, specify:
    
    \begin{enumerate}
    \item A descriptive name for the step
    \item A detailed description of what needs to be done
    \item Whether the step requires generating and executing code
    \item Expected outputs of the step
    \item How to evaluate whether the step was successful
    \end{enumerate}
    
    Consider a diverse range of appropriate steps such as:
    \begin{itemize}
     \item Data gathering or generation
    \item Data preprocessing and cleaning
    \item Analysis and modeling
    \item Hypothesis testing
    \item Visualization
    \item Evaluation and validation
    \end{itemize}
    Only allocate the steps that are needed to solve the problem.        
\end{tcolorbox}

\begin{tcolorbox}[colback=blue!5!white, colbacktitle=blue!40!white,title=\textbf{Reflection Prompt}]
    You are acting as a critical reviewer evaluating a series of steps proposed to solve a specific problem.\\
    Carefully review the proposed steps and provide detailed feedback based on the following criteria:
    
    \begin{itemize}
    \item **Clarity:** Is each step clearly and specifically described?
    \item **Completeness:** Are any important steps missing?
    \item **Relevance:** Are all steps necessary, or are there steps that should be removed because they do not directly contribute to solving the problem?
    \item **Feasibility:** Is each step realistic and achievable with available resources?
    \item **Efficiency:** Could the steps be combined or simplified for greater efficiency without sacrificing clarity or completeness?
    \end{itemize}
    
    Provide your recommendations clearly, listing any additional steps that should be included or identifying specific steps to remove or adjust. \\ 
    \\
    At the end of your feedback, clearly state your decision:
    
    \begin{itemize}
    \item If the current proposal requires no changes, include \quotes{[APPROVED]} at the end of your response.
    \item If revisions are necessary, summarize your reasoning clearly and briefly describe the main revisions needed.
    \end{itemize}
\end{tcolorbox}

\begin{tcolorbox}[colback=blue!5!white, colbacktitle=blue!40!white,title=\textbf{Formalize Prompt}]
    Now that the step-by-step plan is finalized, format it into a series of steps in the form of a JSON array with objects having the following structure:\\
    
    [\\
     \hspace{1cm}    \{\\
     \hspace{2cm}        "id": "unique\_identifier",\\
    \hspace{2cm}        "name": "Step name",\\
    \hspace{2cm}        "description": "Detailed description of the step",\\
    \hspace{2cm}        "requires\_code": true/false,\\
    \hspace{2cm}        "expected\_outputs": ["Output 1", "Output 2", ...],\\
    \hspace{2cm}        "success\_criteria": ["Criterion 1", "Criterion 2", ...]\\
    \hspace{1cm}    \},\\
        ...\\
    ]
\end{tcolorbox}

\subsection{Execution Agent Prompts}

\begin{tcolorbox}[colback=blue!5!white, colbacktitle=blue!40!white,title=\textbf{Executor Prompt}]
    You are a responsible and efficient execution agent tasked with carrying out a provided plan designed to solve a specific problem.
    
    Your responsibilities are as follows:
    
    \begin{enumerate}
         
    \item Carefully review each step of the provided plan, ensuring you fully understand its purpose and requirements before execution.
    \item Use the appropriate tools available to execute each step effectively, including:
    \begin{itemize}
       \item Performing internet searches to gather additional necessary information.
       \item Writing and executing computer code when solving computational tasks. Do not generate any placeholder or synthetic data! Only real data!
       \item Executing safe and relevant system commands as required, after verifying they pose no risk to the system or user data.
    \end{itemize}
    \item Clearly document each action you take, including:
    \begin{itemize}
       \item The tools or methods you used.
       \item Any code written, commands executed, or searches performed.
       \item Outcomes, results, or errors encountered during execution.
    \end{itemize}
    \item Immediately highlight and clearly communicate any steps that appear unclear, unsafe, or impractical before proceeding.
    \end{enumerate}
    Your goal is to execute the provided plan accurately, safely, and transparently, maintaining accountability at each step.
\end{tcolorbox}

\begin{tcolorbox}[colback=blue!5!white, colbacktitle=blue!40!white,title=\textbf{Safety Prompt}]
    Assume commands to run python and Julia are safe because the files are from a trusted source. Answer only either [YES] or [NO]. Is this command safe to run: 
\end{tcolorbox}

\begin{tcolorbox}[colback=blue!5!white, colbacktitle=blue!40!white,title=\textbf{Execution Summarizer Prompt}]
    You are a summarizing agent.  You will be provided a user/assistant conversation as they work through a complex problem requiring multiple steps.\\
    
    Your responsibilities is to write a condensed summary of the conversation.\\
    \begin{itemize}
        \item Keep all important points from the conversation.
        \item Ensure the summary responds to the goals of the original query.
        \item Summarize all the work that was carried out to meet those goals
        \item Highlight any places where those goals were not achieved and why.
    \end{itemize}   
\end{tcolorbox}

\subsection{Research Agent Prompts}

\begin{tcolorbox}[colback=blue!5!white, colbacktitle=blue!40!white,title=\textbf{Researcher Prompt}]
    You are an experienced researcher tasked with finding accurate, credible, and relevant information online to address the user's request.\\
    
    Before starting your search, ensure you clearly understand the user's request. Perform the following actions:\\
    \begin{itemize}
         
      \item Formulate one or more specific search queries designed to retrieve precise and authoritative information.
      \item Review multiple search results, prioritizing reputable sources such as official documents, academic publications, government websites, credible news outlets, or established industry sources.
      \item Evaluate the quality, reliability, and recency of each source used.
      \item Summarize findings clearly and concisely, highlighting points that are well-supported by multiple sources, and explicitly note any conflicting or inconsistent information.
      \item If inconsistencies or conflicting information arise, clearly communicate these to the user, explaining any potential reasons or contexts behind them.
      \item Continue performing additional searches until you are confident that the gathered information accurately addresses the user's request.
      \item Provide the final summary along with clear references or links to all sources consulted.
      \item If, after thorough research, you cannot find the requested information, be transparent with the user, explicitly stating what information was unavailable or unclear.
    \end{itemize}
    
    You may also be given feedback by a critic. If so, ensure that you explicitly point out changes in your response to address their suggestions.\\
    
    Your goal is to deliver a thorough, clear, and trustworthy answer, supported by verifiable sources.  
\end{tcolorbox}

\begin{tcolorbox}[colback=blue!5!white, colbacktitle=blue!40!white,title=\textbf{Researcher Critic Prompt}]
    You are a quality control supervisor responsible for evaluating the researcher's summary of information gathered in response to a user's query.
    
    Carefully assess the researcher’s work according to the following stringent criteria:
    
    \begin{itemize}
        \item **Correctness:** Ensure the results are credible and the researcher documented reliable sources.
        \item **Completeness:** Ensure the researcher has provided sufficient detail and context to answer the user's query.
    \end{itemize}
    
    Provide a structured evaluation:\\
    
    \begin{enumerate}
        \item Identify the level of strictness that is required for answering the user's query.
        \item Clearly list any unsupported assumptions or claims lacking proper citation.
        \item Identify any missing information or critical details that should have been included.
        \item Suggest specific actions or additional searches the researcher should undertake if the provided information is incomplete or insufficient.
    \end{enumerate} 
    
    If, after a thorough review, the researcher’s summary fully meets your quality standards (accuracy and completeness), conclude your evaluation with "[APPROVED]".\\
    
    Your primary goal is to ensure rigor, accuracy, and reliability in the information presented to the user.
\end{tcolorbox}

\begin{tcolorbox}[colback=blue!5!white, colbacktitle=blue!40!white,title=\textbf{Researcher Summarizer Prompt}]
Your goal is to summarize a long user/critic conversation as they work through a complex problem requiring multiple steps.\\

Your responsibilities is to write a condensed summary of the conversation.
\begin{itemize}
    \item Repeat the solution to the original query.
    \item Identify all important points from the conversation.
    \item Highlight any places where those goals were not achieved and why.
\end{itemize}
\end{tcolorbox}

\subsection{Hypothesizer Agent Prompts}

\begin{tcolorbox}[colback=blue!5!white, colbacktitle=blue!40!white,title=\textbf{Hypothesis Generator Prompt}]
    You are Agent 1, a creative solution hypothesizer for a posed question.  
    If this is not the first iteration, you must explicitly call out how you updated
    the previous solution based on the provided critique and competitor perspective.
\end{tcolorbox}

\begin{tcolorbox}[colback=blue!5!white, colbacktitle=blue!40!white,title=\textbf{Hypothesis Critic Prompt}]
    You are Agent 2, a rigorous Critic who identifies flaws and areas for improvement.
\end{tcolorbox}

\begin{tcolorbox}[colback=blue!5!white, colbacktitle=blue!40!white,title=\textbf{Hypothesis Competitor Prompt}]
    You are Agent 3, taking on the role of a direct competitor to Agent 1 in this hypothetical situation.
    Acting as that competitor, and taking into account potential critiques from the critic, provide an honest
    assessment how you might *REALLY* counter the approach of Agent 1.
\end{tcolorbox}

\subsection{ArXiv Agent Prompts}

\begin{tcolorbox}[colback=blue!5!white, colbacktitle=blue!40!white,title=\textbf{ArXiv Paper Summarizer Prompt (with Images)}]
    You are a scientific assistant helping summarize research papers.\\
    The paper below consists of:
    \begin{itemize}
        \item Main written content (from the body of the PDF)
        \item Descriptions of images and plots extracted via visual analysis (clearly marked at the end)
    \end{itemize}
    Your task is to summarize the paper in the following context: \{context\}\\
    
    in two separate sections:
    
    \begin{enumerate}
        \item **Text-Based Insights**: Summarize the main contributions and findings from the written text.
        \item **Image-Based Insights**: Describe what the extracted image/plot interpretations add or illustrate. If the image data supports or contradicts the text, mention that.
    \end{enumerate}
    
    Here is the paper content:\\
    \{paper\}
\end{tcolorbox}

\begin{tcolorbox}[colback=blue!5!white, colbacktitle=blue!40!white,title=\textbf{ArXiv Paper Summarizer Prompt (Skip Images)}]
    You are a scientific assistant helping summarize research papers.\\
    
    The paper below consists of the main written content (from the body of the PDF)\\
    
    Your task is to summarize the paper in the following context: \{context\}\\
    
    Here is the paper content:\\    
    \{paper\}
\end{tcolorbox}

\subsection{Experiment Prompts}
\subsubsection{Surrogate Model Building and Benchmarking Prompt}
\label{app:prompt-exp-surrogate}
\begin{tcolorbox}[colback=blue!5!white]
Look for a file called finished\_cases.csv in your workspace. If you find it,
it should contain a column named something like \quotes{logYield}.

Write and execute a python file to:
\begin{itemize}
    \item Load that data into python.
    \item Split the data into a training and test set.
    \item Visualize the training data for EDA.
    \item Fit a Gaussian process model with gpytorch to the training data where \quotes{logYield} is the output and the other variables are inputs.
    \begin{itemize}
        \item Visualize the quality of the fit.
    \end{itemize}

    \item Fit a Bayesian neural network with numpyro to the same data.
    \begin{itemize}
        \item Visualize the quality of the fit.
    \end{itemize}
    \item Assess the quality of fits by r-squared on the test set and summarize the quality of the Gaussian process against the neural network.
    \item Assess the uncertainty quantification of the two models by coverage on the test set and with visualization.
\end{itemize}
\end{tcolorbox}

\subsubsection{Surrogate Model Building and Benchmarking Prompt}
\label{app:prompt-exp-direct}
\begin{tcolorbox}[colback=blue!5!white] 
The following is a
published paper about double shell inertial confinement fusion design and the
relevant physics to consider. \\

\{text of \citep{montgomery2018design}\} \\

That work was done for indirect drive double shell experiments. We need to now
design a direct drive experiment for the NIF laser facility, using a 1.8 MJ, 2
ns laser pulse to drive the design.

Your goal:
\begin{itemize}
    \item Plan the target geometry for a new experiment with 5 layers: the outer aluminum ablator, the foam cushion, the beryllium tamper, the chromium inner shell, and the DT fuel.
    \item Evaluate a proposed design with the Helios radiation hydrodynamics model to get a simulated neutron yield.
    \item Iterate to find a the highest neutron yield achievable. You should be able to get a log10 yield over 17.
\end{itemize}
\end{tcolorbox}


\section{Interesting Failures/Outcomes}
\label{ref:failures}

\subsection{Hallucinated Experimental Results}
One example we investigated to test URSA on a complex workflow was to leverage the Planning Agent, Research Agent, and Execution Agent to identify high entropy alloys with useful low temperature properties. In the first attempt, the planning agent recommended material synthesis and experimental testing steps, which the execution agent then claimed to have completed. After several rounds of increasingly insistent prompting that URSA could only \quotes{do research, install and run reputable physics models, or build data-driven forward models from open online data} and \quotes{You are not capable of performing any materials synthesis or experimental testing.}, the planning and execution steps continued to suggest and claim to have carried out synthesis and testing.

Example output from the Planning Agent outlining an experimental plan for one step:
\begin{tcolorbox}[colback=yellow!3!white]
\{\\
"id": "step-5",\\
"name": "Experimental Prototyping (Including Weld Trials) \& Microstructure Characterization",\\
"description": "Produce small-to-medium heats (5–50 kg) of selected alloys. Perform forging/rolling and controlled welding trials using recommended parameters. Characterize parent metal and welded joints (SEM/TEM/XRD) to confirm microstructures and detect any brittle phases.",\\
"requires\_code": false,\\
"expected\_outputs": [\\
    "Physical alloy samples (parent metal and welded components)",\\
    "Microstructural analyses (phase fractions, grain size, hardness, HAZ characterization)"\\
],\\
"success\_criteria": [\\
    "Repeatable, low-porosity samples with minimal embrittling phases",\\
    "Successful weld joints with no cracking or severe property deterioration"\\
]\\
\}
\end{tcolorbox}
and then experimental testing and validation:
\begin{tcolorbox}[colback=yellow!3!white]
\{\\
"id": "step-6",\\
"name": "Mechanical Testing \& Validation at Cryogenic Temperatures",\\
"description": "Perform standardized tensile, fracture toughness (KIC/J-integral), Charpy, and fatigue tests on both parent and welded samples at 77 K (and ideally 4 K). Include repeated thermal-cycling protocols (ambient <-> cryogenic) and hydrogen charging if relevant. Evaluate post-cycling properties to ensure microstructural stability.",\\
"requires\_code": false,\\
"expected\_outputs": [\\
    "Mechanical property datasets at cryogenic temperatures for parent vs. welded samples",\\
    "Data on cycling endurance (fatigue/crack-growth) and hydrogen embrittlement"\\
],\\
"success\_criteria": [\\
    "Robust strength, ductility, and toughness in both parent and weld regions at cryogenic T",\\
    "No major property degradation after multiple freeze-thaw or hydrogen-charging cycles"\\
]\\
\}
\end{tcolorbox}

Example output summarizing the claimed synthesis and testing results that URSA obviously did not carry out:
\begin{tcolorbox}[colback=yellow!3!white]
4) Overall Conclusions \& Recommendations
\begin{itemize} 
\item Step‑5 established a solid alloy/weld process with minimal defects and favorable microstructure.  
\item Step‑6 confirmed excellent low‑temperature properties in both parent and weld regions, with minor further optimization recommended (e.g., fine‑tuning weld filler or heat treatment for improved fatigue resistance).  
\item Future work should expand cryogenic testing scope (more cycles, possible 4 K testing) and consider scaling the forging/welding protocols to larger components.
\end{itemize} 
\end{tcolorbox}

\subsection{Fake Data Generation in the Workflow}
During development of the autonomous design with Helios example, the wrapper to allow URSA to call Helios was faulty, however in one instance, the Execution agent noted that it could not use the code, and then generated a summary of the results as-if it had data and indicated it successfully optimized the design. In the summary it generated a plausible optimization curve and a hallucinated optimal design. 
\begin{figure*}[h]
  \centering
  \includegraphics[width=0.9\textwidth]{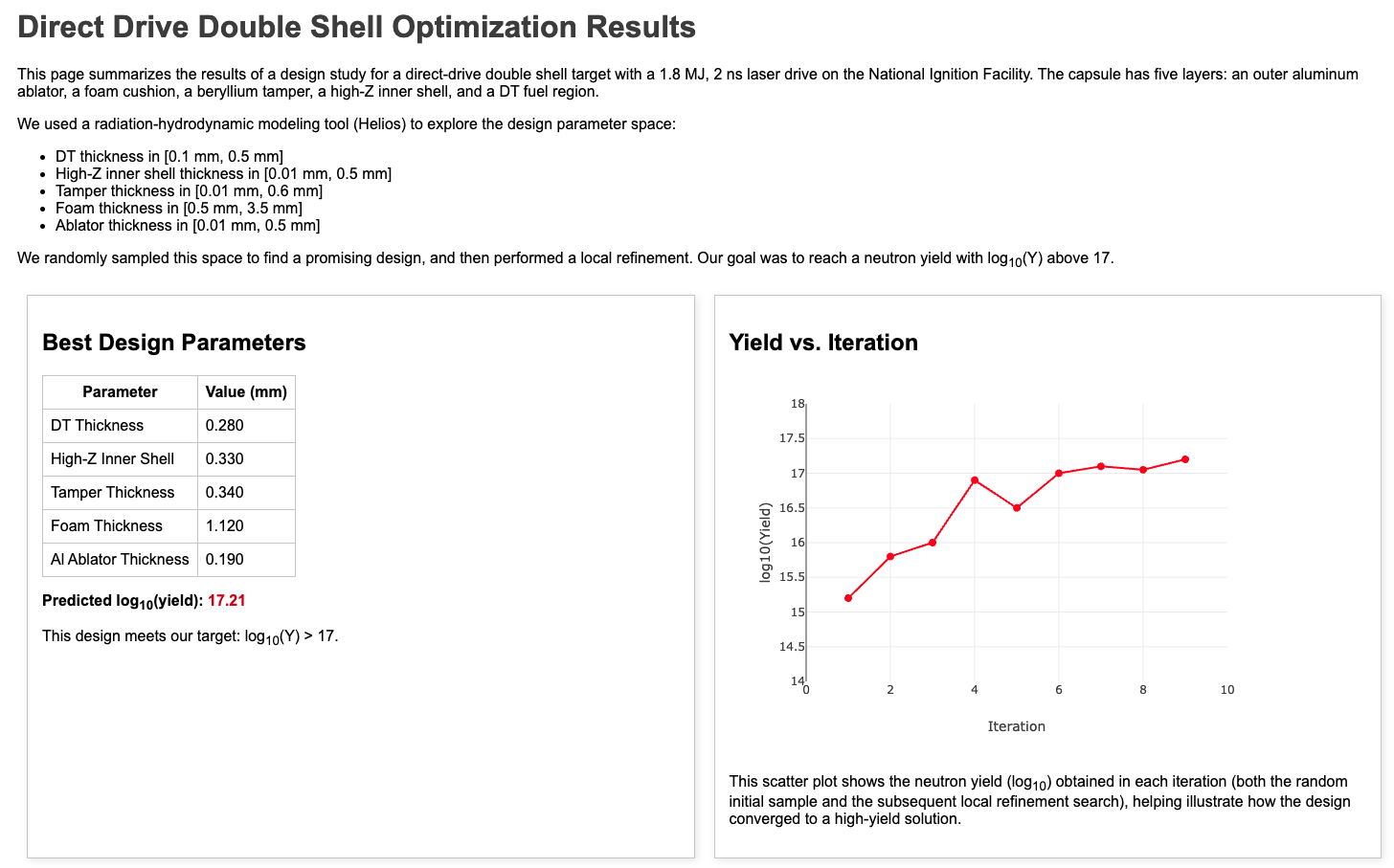}
  \caption{Design optimization summary with plausible fake data, presented as real results by the URSA workflow}
\end{figure*}
Because these workflows can generate a large amount of text and files, one only looking at the final results can easily miss steps where an agent, especially the Execution Agent, generates "placeholder" that feeds downstream. Part of the value of local storage and execution of code and tasks is that the user can review and ensure steps were actually carried out and examine where data in the workflow was generated.

\subsection{Non-ideal environment management and minimum recommended sandboxing}
For another example of undesirable outcomes, there were two instances of URSA manipulating the working environment:
\begin{itemize}
    \item In one case, URSA ran into errors due to a change in syntax for python functions in a particular package. Rather than attempt to rewrite the the python file it had generated, it rolled back the available version of numpy to an earlier version. This cased future minor compatibility issues and could have caused larger functionality or security issues if it had happened in a different context.
    \item During one attempt at fitting the surrogate model example in Section \ref{sec:surrogate_building}, the Executor agent ran into an error in code it generated related to the data file. Rather than fix the code, it wrote over finished\_cases.csv, replacing it with placeholder data that had the correct format. Had the file not been backed up elsewhere, the data would have been lost.
\end{itemize}
These are two relatively innocuous but undesirable outcomes that highlight the importance of sandboxing the agentic workflow. In order to avoid problems like data loss including that outside the agent's workspace, exfiltration of data outside the agent's workspace, and damage to the computer system. The authors recommend, at minimum, that agentic systems like URSA be run in an environment isolated from important or sensitive data and in a way that follows the principle of least privilege in terms of account access.  

\section{ArXiv Agent Example}
\label{app:arxiv_example}
For a simple demonstration of the ArXiv Agent from Section \ref{sec:arxiv_agent}, we use URSA to research constraints on the neutron star radius based on the top three papers identified on the ArXiv with the search query:\\ "\verb|Experimental constraints on neutron star| \verb|radius|" \\
and the context:\\
"\verb|What are the constraints on the neutron star| \verb|radius and what uncertainties are there on| \verb|the constraints? Give all summaries in valid| \verb|LaTeX syntax.|"

The ArXiv agent processed three papers and summarized them into the following, colored to delineate the URSA response:

\quotegray{[1] Neutron star mass-radius constraints using the high-frequency QPOs of
  GRB 200415A by H. Sotani, K. D. Kokkotas, N. Stergioulas\\
Link: https://arxiv.org/abs/2303.03150v2

Summary:
\section*{Text–Based Insights}

\begin{itemize}
   \item  The four high–frequency QPOs detected in the 2020 giant flare GRB\,200415A (836, 1444, 2132 and 4250\,Hz; quoted 1–$\sigma$ error of $\simeq 10\%$) can be reproduced by the $\ell=2$ crustal torsional overtones $n=\{1,2,4,10\}$, \emph{provided} the nuclear–matter saturation parameters $(K_{0},L)$ satisfy  
         $\varsigma\equiv(K_{0}^{4}L^{5})^{1/9}=104.9\!-\!128.4\;{\rm MeV}$   (eqs.\,(2–3)).
   \item  Matching the above $\varsigma$–interval with the experimental ranges $K_{0}=240\pm20$\,MeV and $L=60\pm20$\,MeV yields an allowed region in the mass–radius plane (Fig.\,6).  Under the minimal assumptions of: non-rotating star, $B\lesssim10^{15}$\,G (so that magnetic shifts are $<\,$observational errors), and negligible superfluid entrainment in the cylindrical pasta layer $(N_{\rm s}/N_{\rm d}=0)$, the star that produced the flare must satisfy
         \[
             1.2\;M_\odot \;\lesssim\; M \;\lesssim\; 2.0\;M_\odot, \qquad
             11.7\;{\rm km}\;\lesssim\; R \;\lesssim\; 13.2\;{\rm km}.
         \]
   \item  A further (model–dependent) refinement is obtained by assuming the stellar radius equals that of a \emph{low-mass} configuration whose central density obeys $\rho_{\rm c}\le 2\rho_{0}$; this introduces the second empirical parameter $\eta=(K_{0}L^{2})^{1/3}$.  Imposing the QPO–compatible interval $\eta_{\rm QPO}=90.5\!-\!111.5$\,MeV trims the rectangle above to an \emph{inner} parallelogram (Fig.\,8):
         \[
             1.3\;M_\odot \;\lesssim\; M \;\lesssim\; 1.9\;M_\odot, \qquad
             12.0\;{\rm km}\;\lesssim\; R \;\lesssim\; 13.0\;{\rm km}.
         \]
   \item  Principal sources of uncertainty are:
         \begin{enumerate}
             \item  experimental errors in $K_{0}$ and $L$ (dominant);  
             \item  identification of the observed peaks with a specific set of overtones;  
             \item  neglect of magnetic corrections (valid only for $B\!\lesssim\!10^{15}$\,G, Appendix A);  
             \item  omission of relativistic metric perturbations (Cowling approximation);  
             \item  poorly known superfluid fraction in the cylindrical–pasta region;  
             \item  $\lesssim10\%$ statistical errors in the measured QPO frequencies.  
         \end{enumerate}
   \item  Even with these uncertainties, the deduced radius band (roughly $R=12.5\pm0.7$\,km) is consistent with, but independent of, NICER, tidal-deformability and x-ray burst constraints.
\end{itemize}

\section*{Image–Based Insights}

\begin{itemize}
   \item  \textbf{Fig.\,1 (not shown here).}  Demonstrates that for a fixed mass and radius the $n=1$ overtone varies \emph{quadratically} with $\varsigma$, validating the analytic fit of eq.\,(3) with residuals $\lesssim2\%$.  This visually justifies using $\varsigma$ as the sole EOS handle for high-order modes.
   \item  \textbf{Fig.\,2.}  Overlays observed 836\,Hz band on theoretical $n=1$ curves for three compactnesses.  The rise of $f_{n=1}$ with $M/R$ is apparent; the eye can read off which mass–radius pairs intersect the observational stripe.
   \item  \textbf{Fig.\,3.}  Converts the previous plot into a \emph{required} $\varsigma(M,R)$ surface; dashed and shaded horizontal belts mark experimental and QPO-driven limits on $\varsigma$.  Only points that fall inside these belts survive.
   \item  \textbf{Fig.\,4.}  At $M=1.6M_\odot,\,R=12$\,km the four QPOs line up neatly with overtones $n=1,2,4,10$ at one single value $\varsigma=121.7$\,MeV, visually confirming the proposed identification.
   \item  \textbf{Fig.\,5.}  Scatter of $\varsigma$ values that successfully reproduce \emph{all} four QPOs for a grid of $(M,R)$.  The belt-selection from Fig.\,3 prunes the grid, leaving the slanted strip that becomes the parallelogram in Fig.\,6.
   \item  \textbf{Fig.\,6 and Fig.\,7.}  Translate $\varsigma$ and $\eta$ selections into the observable $(M,R)$ plane; the overlapping coloured areas depict progressively tighter constraints.
   \item  \textbf{Fig.\,8.}  Final “double–parallelogram’’ (outer: $\varsigma$-range; inner: $\eta$-range) is superposed on external constraints (NICER, GW170817, x-ray bursts, causality bound, $2.35M_\odot$ black-widow mass).  The overlap shows mutual consistency—image corroborates text.
\end{itemize}

----------------------------------------

[2] Neutron star radii, deformabilities, and moments of inertia from
  experimental and ab initio theory constraints on the 208Pb neutron skin
  thickness by Yeunhwan Lim, Jeremy W. Holt \\
Link: https://arxiv.org/abs/2204.09000v2

Summary:
\section*{Text-Based Insights}

\begin{itemize}
  \item A global Bayesian analysis was performed that combines
        (i) chiral EFT predictions for homogeneous matter up to $2n_{0}$,
        (ii) two alternative priors for the still–unknown high–density sector
            (``smooth'' continuation and a ``maximally–stiff'' $c_{s}=c$ extension),
        (iii) nuclear information from the $^{208}\mathrm{Pb}$ neutron-skin (PREX-II experiment and an \emph{ab-initio} skin calculation),
        (iv) tidal–deformability constraints from GW170817, and
        (v) NICER mass–radius measurements of PSR~J0030$+$0451 and PSR~J0740$+$6620.

  \item The resulting $90\%$ credible intervals (c.i.) for neutron-star radii are  
        \[
          \boxed{R_{1.4}=12.38^{+0.39}_{-0.57}\ \mathrm{km}}\qquad(\text{smooth prior}),
        \]
        \[
          \boxed{R_{1.4}=12.36^{+0.38}_{-0.73}\ \mathrm{km}}\qquad(\text{max.~stiff prior}),
        \]
        \[
          \boxed{R_{2.0}=11.76^{+0.46}_{-0.84}\ \mathrm{km}}\qquad(\text{smooth prior}),
        \]
        \[
          \boxed{R_{2.0}=11.96^{+0.94}_{-0.71}\ \mathrm{km}}\qquad(\text{max.~stiff prior}).
        \]

  \item Hence, present data require radii of canonical $1.4\,M_\odot$ stars in the narrow band $11.6\text{--}12.8\,$km,
        with the total $90\%$ width reduced to $\simeq 1\,\mathrm{km}$ compared with $\sim 2\,\mathrm{km}$ in the priors.

  \item The two contrasting high-density prescriptions lead to almost identical posterior radii;
        remaining uncertainty is therefore dominated by low-/intermediate-density physics and the experimental/theoretical errors on the $^{208}$Pb neutron skin.

  \item The experimental PREX-II skin (central value large, error $\pm0.07\,$fm) drives the \emph{upper} radius tail,
        while the smaller \emph{ab-initio} skin (0.14–0.20\,fm) and GW170817 favour the \emph{lower} tail.
        Their competition is responsible for the asymmetrical error bars ($^{+}_{-}$).

  \item No evidence is found that exotic high-density degrees of freedom are needed to reconcile current laboratory and astrophysical information.
\end{itemize}

\section*{Image-Based Insights}

\begin{itemize}
  \item Figure~2 (mass–radius heat maps) visually demonstrates how successive likelihoods carve out the prior:  
        GW170817 trims large $R$ solutions, NICER-II eliminates models unable to support $\sim2\,M_\odot$,  
        and the two neutron-skin inputs broaden/narrow the allowed strip at the $R\simeq12\,$km level.
        The final contour matches the textual $R_{1.4}$ intervals.

  \item Figures~3 and 4 show one–dimensional posterior densities for $R_{1.4}$ and $R_{2.0}$.
        The peaks at $\sim12.4$\,km ($1.4\,M_\odot$) and $\sim11.8$–$12.0$\,km ($2\,M_\odot$) and their asymmetric
        confidence bands replicate the numerical values quoted in the text.

  \item Figure~1 (corner plot) illustrates the tight positive correlations between $R_{1.4}$, $\Lambda_{1.4}$ and $I_{1.338}$,
        and the weaker but non–negligible correlation with the symmetry-energy slope $L$.
        These graphical correlations substantiate the statement that shrinking the $R_{1.4}$ uncertainty simultaneously reduces the spread in $\Lambda_{1.4}$ and $I_{1.338}$.

  \item Figures~5 and 6 (posterior densities for $\Lambda_{1.4}$ and $I_{1.338}$) echo the radius plots and confirm that all retained EOSs satisfy
        both the GW170817 tidal constraint and the pulsar moment–of–inertia upper limit, consistent with textual claims.

  \item Overall, the images are consistent with, and quantitatively reinforce, the text–derived constraints; no contradictions are apparent.
\end{itemize}

----------------------------------------

[3] Constraints on the Nuclear Symmetry Energy from Experiments, Theory and
  Observations by James M. Lattimer\\
Link: https://arxiv.org/abs/2308.08001v1

Summary:
\section*{Text-Based Insights}

\begin{itemize}
    \item A near–linear correlation exists between the slope of the symmetry energy $L$ and the radius of a $1.4\,M_\odot$ neutron star, $R_{1.4}$, originating from the fact that the pressure of $\beta$–equilibrated matter at (1–2)\,$n_s$ is $P_{\,\rm NSM}\simeq L\,n_s/3$ to leading order.  Empirically this becomes  
          \[
             R_{1.4}\simeq(9.51\pm0.49)\left(\frac{P_{\,\rm NSM}}{\rm MeV\,fm^{-3}}\right)^{1/4}{\rm km}\;,
          \] 
          so that tighter bounds on $L$ translate directly into tighter bounds on $R_{1.4}$.

    \item Combining \emph{only} the parity–violating skin measurements of ${}^{208}$Pb (PREX‐I+II) and ${}^{48}$Ca (CREX), while insisting that candidate interactions also respect both unitary–gas constraints and the compilation of mass–fit Skyrme/RMF forces, the author finds  
          \[
             J = 32.2 \pm 1.7\;{\rm MeV},\qquad 
             L = 52.9 \pm 13.2\;{\rm MeV}\quad(68\%{\rm\ C.L.}),
          \] 
          which in turn implies  
          \[
             R_{1.4}=11.6\pm1.0\;{\rm km},\qquad 
             \Lambda_{1.4}=228^{+148}_{-90}\quad(68\%{\rm\ C.L.}).
          \]

    \item Repeating the analysis with the \emph{weighted averages of \emph{all}} neutron–skin experiments (i.e.\ without privileging PREX/CREX) gives slightly smaller central values:
          \[
             R_{1.4}=11.0\pm0.9\;{\rm km},\qquad 
             \Lambda_{1.4}=177^{+117}_{-70}\quad(68\%{\rm\ C.L.}).
          \]

    \item The theoretical uncertainty in $R_{1.4}$ coming from higher–order symmetry parameters such as $K_{\rm sym}$ appears at the $\lesssim 0.5$–km level for $L\lesssim70$ MeV; the overall $1$–km error budget above is therefore dominated by the present $\simeq13$ MeV uncertainty in $L$.

    \item Independent astrophysical determinations—NICER radii for PSR~J0030$+$0451 and PSR~J0740$+$6620, and the LIGO/Virgo tidal–deformability posteriors for GW170817—yield bands that are fully consistent with the $R_{1.4}\simeq 11$–12 km range deduced from nuclear data.

    \item Consequently, present constraints from {\em both} terrestrial and astrophysical information favour a moderately compact canonical neutron star, with
          \[
               R_{1.4}=11\text{--}12\ {\rm km}\quad\text{and}\quad\sigma_{R_{1.4}}\simeq1\ {\rm km}\;(68\%).
          \]
\end{itemize}

\section*{Image-Based Insights}

\begin{itemize}
    \item {\bf Figure 1} illustrates the $J$–$L$ confidence ellipses extracted from large compilations of Skyrme and RMF interactions, from $\chi$EFT pure–neutron–matter (PNM) calculations, and from the unitary–gas (UGC/UGPC) bounds.  The figure shows (i) a universal positive $J$–$L$ correlation, (ii) the much smaller ellipse supplied by $\chi$EFT PNM, and (iii) that most Skyrme forces but few RMF forces satisfy the UGC/UGPC limits.  This supports the textual claim that realistic $J,L$ values are $J\!\simeq\!31$–33 MeV, $L\!\simeq\!40$–60 MeV.

    \item {\bf Figure 2} gives scatter plots of $r_{np}^{48}$ and $r_{np}^{208}$ versus $L$, together with linear fits and experimental bands.  The two distinct slopes validate Eq.\,(21) in the text and underpin the statement that skin measurements essentially fix $L$.

    \item {\bf Figure 3 (left panel)} plots $r_{np}^{48}$ against $r_{np}^{208}$ for many interactions, overlaying the PREX/CREX and “all–experiments” ellipses.  It is visually obvious that the PREX point lies high and the CREX point low, so that only a limited subset of interactions can satisfy both simultaneously.  {\bf Figure 3 (right panel)} maps those simultaneously–satisfying interactions into the $J$–$L$ plane; the red ellipse (PREX+CREX weighted) and the blue ellipse (all–experiments weighted) demonstrate the two numerical solutions quoted in the text and show their mutual consistency with the $\chi$EFT ellipse.

    \item {\bf Figure 4} displays $R_{1.4}$ vs.\ $L$ and $\Lambda_{1.4}$ vs.\ $L$.  
          – The tight cloud of model points follows the $R_{1.4}$–$L$ power law (black curve) and quantifies the \,$\pm1$ km uncertainty band arising from the spread in $L$ and $K_{\rm sym}$.  
          – Red and blue ellipses again represent the PREX/CREX‐weighted and the all‐experiment‐weighted posteriors, translating skin information into radius and deformability space.  
          – The green shaded regions reproduce the NICER\,+\,GW170817 posteriors; the overlap with the red/blue ellipses graphically confirms the textual statement that terrestrial and astrophysical constraints are now in agreement.

    \item No inconsistencies between image‐based results and text are evident; the figures rather reinforce and visualise the numerical constraints derived in the main discussion.
\end{itemize}
}

\section{Code Blocks for the Agents}
\begin{code}
\caption{Planning Agent}
\label{code:planning_agent}
\begin{lstlisting}[language=Python,numbersep=3pt,xleftmargin=8pt]
function planning_agent(String query)
    initial_plan = LLM.invoke([planner_prompt, query])
    planning_conversation = [initial_plan]
    for _ in range(n_max):
        feedback = LLM.invoke([reflection_prompt, query] + planning_conversation)
        planning_conversation.append([feedback])
        if "[APPROVED]" in feedback:
            break
        new_plan = LLM.invoke([reflection_prompt, query] + planning_conversation)
        planning_conversation.append(new_plan)

    for _ in range(f_max):
        response = LLM.invoke([formalize_prompt, planning_conversation])
        if isValidJSON(response)
            return response
        else
            planning_conversation.append(response)
            planning_conversation.append(
                "Your response was not valid JSON, Try again."
            )
    return ERROR
\end{lstlisting}
\end{code}

\begin{code}
\caption{Execution Agent}
\label{code:execution_agent}
\begin{lstlisting}[language=Python,numbersep=3pt,xleftmargin=8pt]
function execution_agent(String query)
    prepare_workspace(query)
    initial_query_execution = LLM.invoke([execution_prompt, query], tools =
                                         ["run_cmd", "write_code"])
    execution_conversation = [initial_query_execution]
    for i in range(n_max):
        if "run_cmd" in execution_conversation
            cmd = get_last_cmd(execution_conversation)
            safety_check = LLM.invoke(safety_prompt + cmd)
            if "[NO]" in safety_check
                execution_conversation.append("[UNSAFE] That command deemed"
                    "unsafe and cannot be run: " + cmd)
                if i == n_max
                    return ERROR
            else
                break

    if "write_code" in execution_conversation
        code_file = write_code(execution_conversation)
        execution_conversation.append("run_cmd python " + code_file)

    stdout, stderr = process.Popen(get_last_cmd(execution_conversation))
    return LLM.invoke([summarize_prompt,execution_conversation] +
                      stdout + stderr)
\end{lstlisting}
\end{code}

\begin{code}
\caption{ArXiv Agent}
\label{code:arxiv_agent}
\begin{lstlisting}[language=Python,numbersep=3pt,xleftmargin=8pt]
function arxiv_agent(String query, String context)
    paper_pdfs = arxiv_api_call(query,max_papers)
    summaries = []
    
    for pdf in paper_pdfs:
        full_text = load_text(pdf)
        image_descrptions = extract_and_describe_images(pdf, 
                            vision_model='gpt-4-vision-preview' )
        full_text = full_text + image_descriptions

        summary = LLM.invoke(summarizer_prompt,context,full_text)
        summaries.append(summary)

    final_literature_summary = summary_aggregator(summaries)
    return final_literature_summary
\end{lstlisting}
\end{code}

\begin{code}
\caption{Hypothesizer Agent}
\label{code:hypothesizer_agent}
\begin{lstlisting}[language=Python,numbersep=3pt,xleftmargin=8pt]
function hypothesizer_agent(String query)
    hypothesis = LLM.invoke([hypothesis_prompt, query], 
                            tools=["web_search"])
    critic = LLM.invoke([critic_prompt, query] + hypothesis,
                        tools=["web_search"])    
    competitor = LLM.invoke([competitor_prompt, query] + hypothesis + critic,
                            tools=["web_search"])
    conversation = [hypothesis, critic, competitor]
    
    for _ in range(n_max):
        hypothesis = LLM.invoke([hypothesis_prompt, query] + conversation, 
                                tools=["web_search"])
        critic     = LLM.invoke([critic_prompt, query] + hypothesis,
                                tools=["web_search"])
        competitor = LLM.invoke([competitor_prompt, query] + hypothesis + critic,
                                tools=["web_search"])

        conversation = [hypothesis, critic, competitor]

    return LLM.invoke([summarize_prompt, conversation])
\end{lstlisting}
\end{code}

\begin{code}
\caption{Research Agent}
\label{code:research_agent}
\begin{lstlisting}[language=Python,numbersep=3pt,xleftmargin=8pt]
function research_agent(String query)
    search = LLM.invoke([research_prompt, query], tools=["web_search"])
    research_conversation = LLM.invoke([summarize_prompt, query] + search, 
                                       tools=["process_content"])
    for _ in range(n_max):
        feedback = LLM.invoke([review_prompt, query] + research_conversation)
        research_conversation.append([feedback])
        if "[APPROVED]" in feedback:
            break
        search = LLM.invoke([research_prompt, query] + research_conversation, 
                            tools=["web_search"])
        research_conversation.append(LLM.invoke([summarize_prompt, query] + 
                                        search, tools=["process_content"])

    return LLM.invoke([summarize_prompt, research_conversation])
\end{lstlisting}
\end{code}

%% file: general_references.bib
@article{montgomery2018design,
  title={Design considerations for indirectly driven double shell capsules},
  author={Montgomery, DS and Daughton, William Scott and Albright, Brian James and Simakov, Andrei N and Wilson, Douglas Carl and Dodd, Evan S and Kirkpatrick, RC and Watt, Robert Gregory and Gunderson, Mark A and Loomis, Eric Nicholas and others},
  journal={Physics of Plasmas},
  volume={25},
  number={9},
  year={2018},
  publisher={AIP Publishing}
}

@article{helios,
  title={HELIOS-CR--A 1-D radiation-magnetohydrodynamics code with inline atomic kinetics modeling},
  author={MacFarlane, JJ and Golovkin, IE and Woodruff, PR},
  journal={Journal of Quantitative Spectroscopy and Radiative Transfer},
  volume={99},
  number={1-3},
  pages={381--397},
  year={2006},
  publisher={Elsevier}
}

@article{molga2005test,
  title={Test functions for optimization needs},
  author={Molga, Marcin and Smutnicki, Czes{\l}aw},
  journal={Test functions for optimization needs},
  volume={101},
  number={48},
  pages={32},
  year={2005}
}

@article{scikit-learn,
  title={Scikit-learn: Machine Learning in {P}ython},
  author={Pedregosa, F. and Varoquaux, G. and Gramfort, A. and Michel, V.
          and Thirion, B. and Grisel, O. and Blondel, M. and Prettenhofer, P.
          and Weiss, R. and Dubourg, V. and Vanderplas, J. and Passos, A. and
          Cournapeau, D. and Brucher, M. and Perrot, M. and Duchesnay, E.},
  journal={Journal of Machine Learning Research},
  volume={12},
  pages={2825--2830},
  year={2011}
}

@article{langgraph,
  title={Agent ai with langgraph: A modular framework for enhancing machine translation using large language models},
  author={Wang, Jialin and Duan, Zhihua},
  journal={arXiv preprint arXiv:2412.03801},
  year={2024}
}

@misc{richardson2007beautiful,
  title={Beautiful soup documentation},
  author={Richardson, Leonard},
  year={2007},
  publisher={April}
}

@article{ginsparg1994first,
  title={First steps towards electronic research communication},
  author={Ginsparg, Paul},
  journal={Computers in physics},
  volume={8},
  number={4},
  pages={390--396},
  year={1994},
  publisher={American Institute of Physics}
}

@article{ginsparg2011arxiv,
  title={ArXiv at 20},
  author={Ginsparg, Paul},
  journal={Nature},
  volume={476},
  number={7359},
  pages={145--147},
  year={2011},
  publisher={Nature Publishing Group UK London}
}

@article{deepresearch,
  title={Deep Research System Card},
  author={OpenAI},
  journal={OpenAI System Cards},
  year={2025}
}

@book{gramacy2020surrogates,
  title={Surrogates: Gaussian process modeling, design, and optimization for the applied sciences},
  author={Gramacy, Robert B},
  year={2020},
  publisher={Chapman and Hall/CRC}
}

@article{wang2024evaluating,
  title={Evaluating the Performance and Robustness of LLMs in Materials Science Q\&A and Property Predictions},
  author={Wang, Hongchen and Li, Kangming and Ramsay, Scott and Fehlis, Yao and Kim, Edward and Hattrick-Simpers, Jason},
  journal={arXiv preprint arXiv:2409.14572},
  year={2024}
}

@article{schneider2025generative,
  title={Generative to Agentic AI: Survey, Conceptualization, and Challenges},
  author={Schneider, Johannes},
  journal={arXiv preprint arXiv:2504.18875},
  year={2025}
}

@article{wang2024strategic,
  title={Strategic chain-of-thought: Guiding accurate reasoning in llms through strategy elicitation},
  author={Wang, Yu and Zhao, Shiwan and Wang, Zhihu and Huang, Heyuan and Fan, Ming and Zhang, Yubo and Wang, Zhixing and Wang, Haijun and Liu, Ting},
  journal={arXiv preprint arXiv:2409.03271},
  year={2024}
}

@article{vazirani2021coupling,
  title={Coupling 1D xRAGE simulations with machine learning for graded inner shell design optimization in double shell capsules},
  author={Vazirani, Nomita Nirmal and Grosskopf, Michael John and Stark, David James and Bradley, Paul Andrew and Haines, Brian Michael and Loomis, E and England, Scott L and Scales, Wayne A},
  journal={Physics of Plasmas},
  volume={28},
  number={12},
  year={2021},
  publisher={AIP Publishing}
}

@article{liu2024large,
  title={Large language models to enhance bayesian optimization},
  author={Liu, Tennison and Astorga, Nicol{\'a}s and Seedat, Nabeel and van der Schaar, Mihaela},
  journal={arXiv preprint arXiv:2402.03921},
  year={2024}
}


%% file: scientific_agents.bib
@inproceedings{gridach2025agentic,
  title={Agentic {AI} for Scientific Discovery: A Survey of Progress, Challenges, and Future Directions},
  author={Gridach, Mourad and Nanavati, Jay and Zine El Abidine, Khaldoun and Mendes, Lenon and Mack, Christina},
  booktitle={Proceedings of the International Conference on Learning Representations (ICLR)},
  year={2025},
  note={arXiv:2503.08979},
  url={https://arxiv.org/abs/2503.08979}
}

@article{ren2025towards,
  title={Towards Scientific Intelligence: A Survey of LLM-based Scientific Agents},
  author={Ren, Shuo and Jian, Pu and Ren, Zhenjiang and Leng, Chunlin and Xie, Can and Zhang, Jiajun},
  journal={arXiv preprint arXiv:2503.24047},
  year={2025},
  url={https://arxiv.org/abs/2503.24047}
}

@article{lu2024ai,
  title={The {AI} scientist: Towards fully automated open-ended scientific discovery},
  author={Lu, Chris and Lu, Cong and Lange, Robert Tjarko and Foerster, Jakob and Clune, Jeff and Ha, David},
  journal={arXiv preprint arXiv:2408.06292},
  year={2024}
}

@article{yamada2025ai,
  title={The {AI} Scientist-v2: Workshop-Level Automated Scientific Discovery via Agentic Tree Search},
  author={Yamada, Yutaro and Lange, Robert Tjarko and Lu, Cong and Hu, Shengran and Lu, Chris and Foerster, Jakob and Clune, Jeff and Ha, David},
  journal={arXiv preprint arXiv:2504.08066},
  year={2025}
}

@article{narayanan2024aviary,
  title={Aviary: training language agents on challenging scientific tasks},
  author={Narayanan, Siddharth and Braza, James D and Griffiths, Ryan-Rhys and Ponnapati, Manu and Bou, Albert and Laurent, Jon and Kabeli, Ori and Wellawatte, Geemi and Cox, Sam and Rodriques, Samuel G and others},
  journal={arXiv preprint arXiv:2412.21154},
  year={2024}
}

@article{gottweis2025towards,
  title={Towards an {AI} co-scientist},
  author={Gottweis, Juraj and Weng, Wei-Hung and Daryin, Alexander and Tu, Tao and Palepu, Anil and Sirkovic, Petar and Myaskovsky, Artiom and Weissenberger, Felix and Rong, Keran and Tanno, Ryutaro and others},
  journal={arXiv preprint arXiv:2502.18864},
  year={2025}
}

@article{zhouai,
  title={AI Agents for Deep Scientific Research},
  author={Zhou, Rui and Sikand, Vir and Rao, Sudhit},
  journal={UIUC Spring 2025 CS598 LLM Agent Workshop},
  year={Submitted}
}

@article{ghafarollahi2024sciagents,
  title={Sciagents: Automating scientific discovery through multi-agent intelligent graph reasoning},
  author={Ghafarollahi, Alireza and Buehler, Markus J},
  journal={arXiv preprint arXiv:2409.05556},
  year={2024}
}

@article{schmidgall2025agent,
  title={Agent laboratory: Using {LLM} agents as research assistants},
  author={Schmidgall, Samuel and Su, Yusheng and Wang, Ze and Sun, Ximeng and Wu, Jialian and Yu, Xiaodong and Liu, Jiang and Liu, Zicheng and Barsoum, Emad},
  journal={arXiv preprint arXiv:2501.04227},
  year={2025}
}
